\documentclass[letterpaper]{article} 
\usepackage{aaai2026}  
\usepackage{times}  
\usepackage{helvet}  
\usepackage{courier}  
\usepackage[hyphens]{url}  
\usepackage{graphicx} 
\usepackage{natbib}  
\usepackage{caption} 
\usepackage{algorithm}
\usepackage{algorithmic}
\usepackage{amsmath}
\usepackage{xcolor}
\usepackage{multirow}
\usepackage{booktabs}
\usepackage{tabularx}
\usepackage{makecell}
\usepackage{amssymb}
\usepackage{amsmath}
\usepackage{newfloat}
\usepackage{listings}
\usepackage[dvipsnames]{xcolor}
\definecolor{custompink}{RGB}{255,105,180} 

\usepackage{newfloat}
\usepackage{listings}
\DeclareCaptionStyle{ruled}{labelfont=normalfont,labelsep=colon,strut=off} 
\floatstyle{ruled}
\newfloat{listing}{tb}{lst}{}
\floatname{listing}{Listing}
\newcommand{\methodname}{VTinker}

\title{\methodname: Guided Flow Upsampling and Texture Mapping \\  for High-Resolution Video Frame Interpolation}
\author{
    Chenyang Wu\textsuperscript{\rm 1}, \quad 
    Jiayi Fu\textsuperscript{\rm 1}, \quad
    Chun-Le Guo\textsuperscript{\rm 1, \rm 2}, \quad 
    Shuhao Han\textsuperscript{\rm 1}, \quad 
    Chongyi Li\textsuperscript{\rm 1, \rm 2} \thanks{Corresponding author.}\\
}
\affiliations{
    \textsuperscript{\rm 1}VCIP, CS, Nankai University \\
    \textsuperscript{\rm 2}NKIARI, Shenzhen Futian \\ 

    \{chenyangwu, jiayifu, hansh\}@mail.nankai.edu.cn,\\
    \{guochunle, lichongyi\}@nankai.edu.cn \\
}

\usepackage{bibentry}

\begin{document}

\maketitle
\begin{abstract}
Due to large pixel movement and high computational cost, estimating the motion of high-resolution frames is challenging.
Thus, most flow-based Video Frame Interpolation (VFI) methods first predict bidirectional flows at low resolution and then use high-magnification upsampling (e.g., bilinear) to obtain the high-resolution ones. 
However, this kind of upsampling strategy may cause blur or mosaic at the flows' edges. 
Additionally, the motion of fine pixels at high resolution cannot be adequately captured in motion estimation at low resolution, which leads to misalignment in task-oriented flows.
With such inaccurate flows, input frames are warped and combined pixel-by-pixel, resulting in ghosting and discontinuities in the interpolated frame. 
In this study, we propose a novel VFI pipeline, {\methodname}, which consists of two core components: guided flow upsampling (GFU) and Texture Mapping. 
After motion estimation at low resolution, GFU introduces input frames as guidance to mitigate detail blurring in bilinear upsampling flows, which makes flows' edges clearer. 
Subsequently, to avoid pixel-level ghosting and discontinuities, Texture Mapping generates an initial interpolated frame, referred to as the intermediate proxy. 
The proxy serves as a cue for selecting clear texture blocks from the input frames, which are then mapped onto the proxy to facilitate producing the final interpolated frame via a reconstruction module.
Extensive experiments demonstrate that {\methodname} achieves state-of-the-art performance in VFI.  
Code is available at: \textcolor{custompink}{\texttt{https://github.com/Wucy0519/VTinker}}.
\end{abstract}

\section{Introduction}
\label{sec:intro}

Recent advancements in imaging technologies and devices have made high-resolution video more widely accessible.
However, due to hardware limitations and network transmission constraints, high-resolution videos are often at a low frame rate.
To improve the frame rate of high-resolution video, video frame interpolation (VFI) that synthesizes intermediate frames between consecutive frames in a video sequence is commonly applied~\cite{reda2022film}.

\begin{figure}[t]
    \centering
    \includegraphics[width=\linewidth]{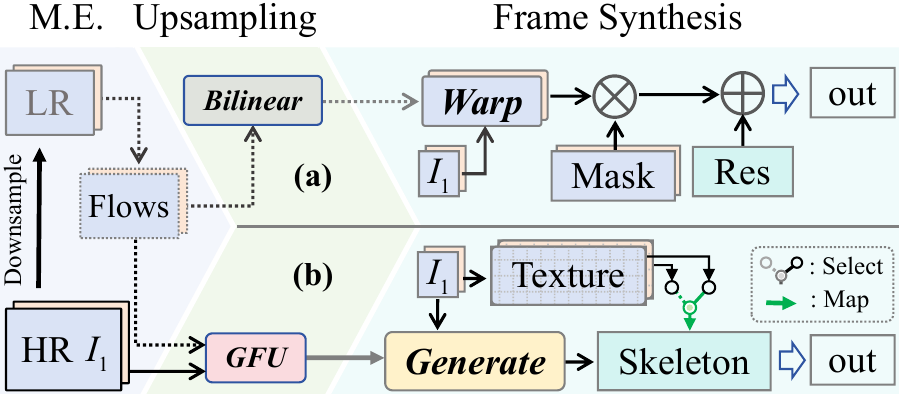}
    \caption{This figure compares {\methodname} with current flow-based method for high-resolution VFI. After Motion Estimation (M.E.), bidirectional flows are obtained at low resolution. \textbf{(a)} The current method employs bilinear upsampling for the high-resolution flows, followed by a pixel-by-pixel synthesis. \textbf{(b)} {\methodname} employs the proposed Guided Flow Upsampling (GFU) and generates the final result through texture mapping. }
    \label{fig:firstfig}
\end{figure}

Benefiting from the advantage of motion estimation, flow-based VFI methods are becoming mainstream. 
As shown in Fig.~\ref{fig:firstfig}, flow-based high-resolution VFI can be divided into three interrelated stages. 
(1) \textit{Motion Estimation (M.E.)}: Given the challenges of larger movements and higher computational demands, motion estimation is performed on the downsampled frames in most high-resolution VFI methods. 
(2) \textit{Upsampling}: To apply low-resolution motion estimation to high-resolution frames, flow-based methods commonly use high-magnification upsampling (e.g., bilinear or adaptive-kernel based) on the low-resolution flow. 
(3) \textit{Frame Synthesis}: Depending on the upsampled flow, the frame synthesis model predicts an intermediate frame between two consecutive frames.

Unfortunately, there are deficiencies in existing flow-based high-resolution VFI pipelines. Specifically, as shown in Fig.~\ref{fig:firstfig}(a), in the context of two consecutive high-resolution frames, ${I_0}$ \& ${I_1}$, the current approach involves frame downsampling and performs flow estimation through motion estimation at a low resolution~\cite{li2023amt, liu2024sparse}.
Subsequently, flow-based methods~\cite{jin2023unified, liu2024sparse} upsample the estimated low-resolution flows using algorithms such as \textbf{Bilinear Upsampling or Adaptive-kernel Based Method} that is called adaptive flow upsampling (AFU, proposed in ASFlow~\cite{Luo2021ASFlowUO} for flow estimation task and used in SGM~\cite{liu2024sparse} for VFI).
%
%
However, in the face of task-oriented flow upsampling during end-to-end training, AFU produces discontinuous boundaries, which also makes it less suitable for flow-based VFI models (such as widespread mosaic-like artifacts in SGM's results, as shown in Figs.~\ref{fig:viscom2k} and ~\ref{fig:viscom4k}).
The more commonly used method is bilinear upsampling. 
However, it generally causes the blurring motion boundaries~\cite{Luo2021ASFlowUO} and further leads to blurring or ghosting.

Additionally, the robust application for the task-oriented flow, which is estimated at low resolution but applied to high resolution, presents a significant challenge.
In the context of motion estimation at low resolution, it is evident that the motion of fine pixels at high resolution is not adequately captured. 
The downsampling operation causes the model to prioritize overall motion and leads to neglecting finer details. 
When flow is upsampled, task-oriented flow encounters difficulties in generalizing to high-resolution applications, which results in biased motion alignment.   

Furthermore, in the frame synthesis stage, the given frames ${I_0}$ \& ${I_1}$ are warped to ${I_t^0}$ \& ${I_t^1}$ with obtained flows, respectively.
They are then merged into an intermediate frame ${{\hat I}_t^{'}}$, which incorporates the predicted mask. 
Then, a network is used to predict a residual term ${Res}$ and add it to the frame ${\hat I_t^{'}}$ to produce the final output ${{\hat I}_t}$ at the pixel level~\cite{jin2023unified, li2023amt}. 
This frame synthesis mechanism, called \textbf{`Mask\&Res’}, produces results at the pixel level with dual-source texture from both ${I_0}$ and ${I_1}$, as shown in Fig.~\ref{fig:firstfig}(a). 
However, the motion estimation is inaccurate due to the blurred boundary and unalignment of the upsampled flows. 
This results in the interpolation generally exhibiting ghosting, blurring, and discontinuities, because it is based on a fusion of the two misaligned warped frames. 

To improve the performance of the high-resolution VFI model and yield clearer results, as shown in Fig.~\ref{fig:firstfig}(b), we propose a novel pipeline, {\methodname}. 
{\methodname} consists of two core components: \textbf{Guided Flow Upsampling (GFU) and Texture Mapping}.
Inspired by UPFlow~\cite{luo2021upflow}, which uses high-resolution information as guidance for flow upsampling in flow estimation task, we propose GFU for task-oriented flow refinement. 
Guided by high-resolution information, the upsampling process using GFU produces sharper flows, aligning more closely with the edges of associated frames. 
GFU helps to provide better high-resolution motion estimation, which improves the quality of the result and reduces artifacts at the boundaries.

In order to reduce ghosting and discontinuities due to the misaligned flows and the `Mask\&Res' mechanism, {\methodname} uses texture mapping to synthesize the interpolated frame. 
Specifically, {\methodname} first generates an intermediate proxy, and then maps it using blocks of regional textures, which are exclusively selected from frame ${I_0}$ or ${I_1}$. 
These continuous block textures, sourced solely from either ${I_0}$ or ${I_1}$, remain unaffected by incorrect warping, resulting in a higher degree of continuity in the texture of ${{\hat I}_t}$.
Finally, the reconstruction module rebuilds the proxy that is mapped by high-quality textures, producing clearer results. 
To ensure that the textures are of high quality, we use the weight-shared reconstruction module to reconstruct the extracted textures and supervise them with input frames.

The \textbf{contributions} of this paper are as follows:

\begin{itemize}
\item To address the boundary errors caused by current upsampling strategies, we propose Guided Flow Upsampling (GFU) inspired by UPFlow~\cite{luo2021upflow}. It improves flow upsampling by using input frames to guide the refinement of task-oriented flows, effectively reducing ghosting along the boundaries in interpolated results. 

\item To resolve flow misalignment resulting from the absence of high-resolution and fine-pixel motion, we propose Texture Mapping. It restores unaligned details using clear textures selected from input frames, improving the continuity in interpolated results. 

\item With our two core components, we propose {\methodname}, a novel VFI pipeline for high-resolution video, achieving state-of-the-art performance in VFI. In particular, for high-resolution VFI, the proposed {\methodname} exhibits impressive performance in the details.
\end{itemize}

\begin{figure*}[t]
    \centering
    \includegraphics[width=\linewidth]{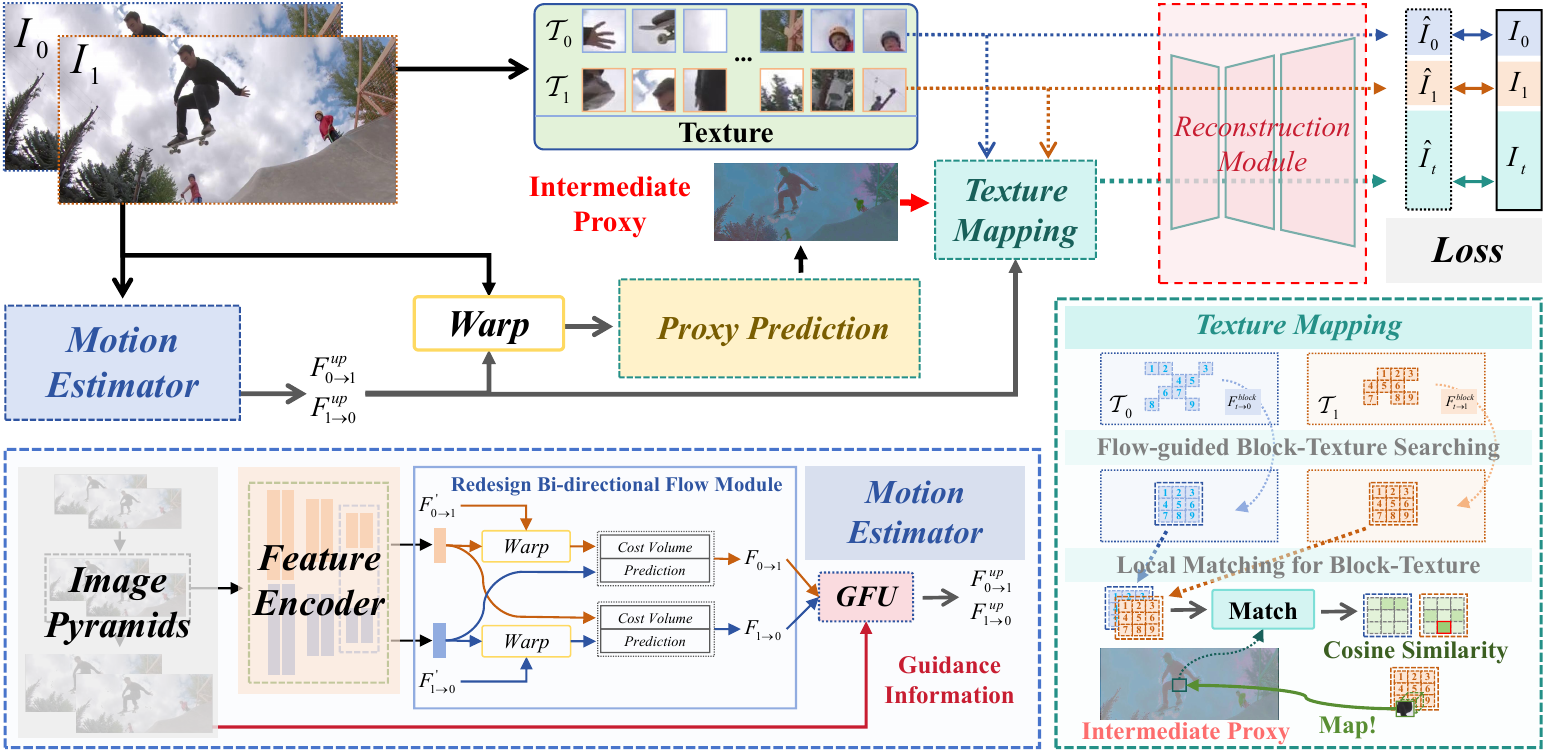}
    \caption{\textbf{Architecture overview of the proposed {\methodname}.} Given two consecutive frames ${I_0}$ \& ${I_1}$, {\methodname} first estimates the bi-directional flows ${F_{0 \to 1}}$ \& ${F_{1 \to 0}}$, which are fused by warping to produce an intermediate proxy. Then, after extracting features of the input frames, {\methodname} divides the features into texture blocks. Through Flow-guided Block-Texture Searching and Local Matching for Block-Texture, texture blocks corresponding to each position of the proxy are selected. Finally, {\methodname} maps these block-textures to the proxy and rebuilds the interpolated frame by a reconstruction module. }
    \label{fig:ppl}
\end{figure*}

\section{Related Work}
\label{sec:relatedwork}

Flow-based VFI~\cite{jin2023unified,li2023amt} is mainstream, almost all flow-based methods first obtain the bidirectional flows through motion estimation and then use the flows to warp the two input frames. 
The final output is obtained by fusing the two warped frames with an estimated Mask and Residual Maps at the pixel level, which is called `Mask\&Res'. 
The texture of the `Mask\&Res' output is pixel-level dual-source and often suffers from ghosting and blurring when motion estimation is inaccurate. 

Facing high-resolution video, VFI models often need to deal with motion over a distance of more than a hundred pixels, making motion estimation particularly challenging~\cite{reda2022film}. 
Some recent methods~\cite{jin2023enhanced, jin2023unified, li2023amt, liu2024sparse} first estimate the flows at low resolution and then warp the original high-resolution frames by the upsampled flows. 
However, the upsampling methods mostly are bilinear, which results in the upsampled flows being misaligned with the images' boundaries, as illustrated in Fig.~\ref{fig:gfu}. 
ASFlow~\cite{Luo2021ASFlowUO} proposes Adaptive Flow Upsampling (AFU), which obtains the optical flow upsampling results by estimating a weighted kernel for each upsampled region separately. 
AFU is considered an effective unsupervised method in the optical flow estimation task. 
However, for VFI, we train the model in an end-to-end manner, which leads to task-oriented flow estimation.
Whether through self-supervision or unsupervision, there is no access to the ground truth of task-oriented flow. 
When the AFU-like upsampling method is used for VFI, the results interpolated by SGM~\cite{liu2024sparse} (shown in Fig.~\ref{fig:viscom4k}) are poorly generalized to high-resolution frame interpolation, resulting in some mosaic-like appearance of results. 
Meanwhile, in Fig.~\ref{fig:xrsy}, ablation experiments demonstrate the discontinuities in the boundary of the flow, which is upsampled by AFU.

\section{Proposed Method: {\methodname}}
\label{sec:method}
As shown in Fig.~\ref{fig:ppl}, given two frames ${I_0}$ \& ${{I_1} \in {\mathbb{R}^{H \times W \times 3}}}$, where ${H}$ and ${W}$ respectively denotes the height and width of frames, our approach aims to obtain the interpolated frame ${\hat I_t}$ according to the time step ${t}$. 
First of all, based on two input frames, {\methodname} initially estimates the bidirectional flows at low resolution, and then upsamples them by GFU. 
Subsequently, {\methodname} generates an intermediate proxy and maps clear textures onto the proxy. 
Finally, {\methodname} obtains the interpolated frame ${{\hat I}_t}$ after being reconstructed by a reconstruction module.

\subsection{Motion Estimation with GFU}
\label{sec:motiones} 
Benefiting from its pyramid recurrent structure, UPR-Net~\cite{jin2023unified} can serve as a lightweight motion estimator for {\methodname}. 
However, the alignment in the UPR-Net~\cite{jin2023unified}'s bi-directional flow module performs in interpolation time step ${t}$, while refines flows ${F_{0 \to 1}}$ and ${F_{1 \to 0}}$. 
This discrepancy could confuse the coordination, which ultimately leads to a reduction in the efficiency of flow estimation. 
To achieve more efficient alignment and more accurate flow estimation, as illustrated in Fig.~\ref{fig:ppl}, we refer to the structure of PWC-Net~\cite{sun2018pwc} and redesign the motion estimator. 

Specifically, the features of ${I_0}$ are warped by predicted flow ${F_{0 \to 1}^{'}}$ and aligned with the features of ${I_1}$, which are used to update flow ${F_{0 \to 1}^{'}}$ and obtain flow ${F_{0 \to 1}}$. The same operation is used for flow ${F_{1 \to 0}^{'}}$, leading to more accurate motion estimation. We train the redesigned UPR-base following the same training setting as UPR-Net. Using the redesigned UPR-base, we obtain low-resolution bi-directional flows ${F_{0 \to 1}}$ and ${F_{1 \to 0}}$, and then we upsample them by GFU.

\begin{figure}[t]
    \centering
    \includegraphics[width=\linewidth, height=0.6\linewidth]{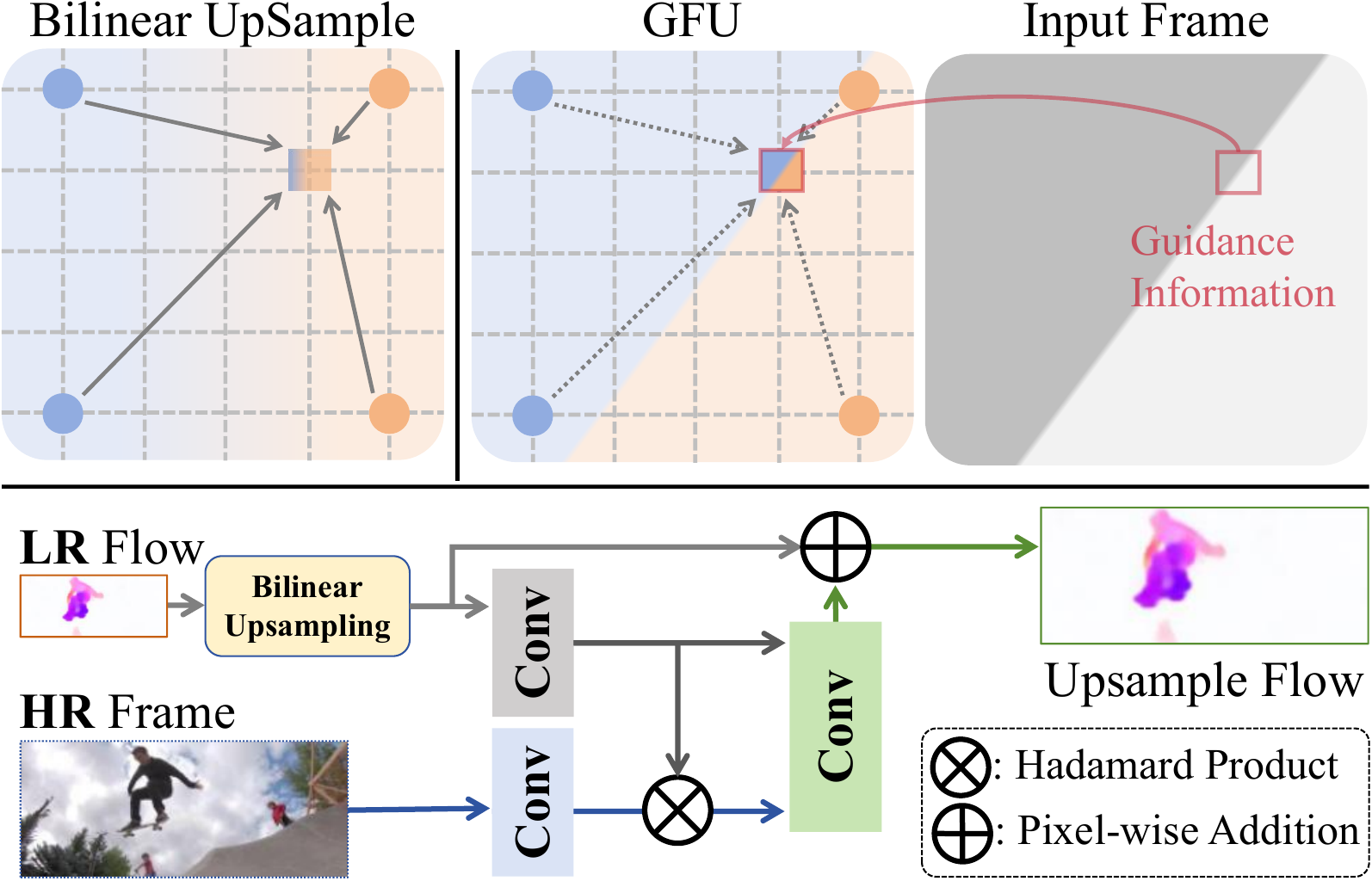}
    \caption{Guided Flow Upsampling (GFU) Module. The difference between bilinear upsampling and GFU is visually shown above the line. The edges of flow upsampled by GFU are more aligned with the input frame than the bilinear. 
    Blue and orange points indicate pixels at low resolution. 
    The framework of GFU is below the line. 
    GFU uses the input frame as guidance information to refine the flow upsampling. Remarkable contrast is shown in Tab.~\ref{tab:xrsy} and Fig.~\ref{fig:xrsy}.
    }
    \label{fig:gfu}
\end{figure}

The boundary of the upsampled flow ${F_{0 \to 1}^{up}}$ should be aligned with the boundary of the frame ${I_0}$.
As shown in Fig.~\ref{fig:gfu}, to address the inherent challenges associated with flow upsampling, inspired by UPFlow~\cite{luo2021upflow}, we propose a simple but effective guided flow upsampling (GFU) module, which integrates high-resolution image information to enhance the quality of the upsampled flow. 
First, GFU uses bilinear flow upsampling to upsample low-resolution flow, producing an initial upsampled flow in which motion edges may appear blurring.
Second, the guided information within the input frame is extracted using convolutional layers. 
Third, GFU employs this guidance information to rectify the blurring details in the bilinearly upsampled flow, enhancing the sharpness of the flow's edges.
Ablation experiments (see Fig.~\ref{fig:xrsy}) demonstrate that GFU offers greater reliability compared to the bilinear upsampling and AFU. 

After upsampled by GFU module, flows ${F_{0 \to 1}^{up}}$ \& ${F_{1 \to 0}^{up}}$ warp frames ${I_0}$ \& ${I_1}$ into ${I_t^0}$ \& ${I_t^1}$, which can be expressed as: 
\begin{equation}
{I_t^0 = {\cal{W}}({I_0},{F_{0 \to {\rm{t}}}^{up}})}, {I_t^1 = {\cal{W}}({I_1},{F_{1 \to {\rm{t}}}^{up}})},
\label{eq:eq0}
\end{equation} 
where ${{F_{0 \to {\rm{t}}}^{up}} = t \times {F_{0 \to 1}^{up}}}$ and ${{F_{1 \to {\rm{t}}}^{up}} = (1-t) \times {F_{1 \to 0}^{up}}}$ denote the estimated flows ${F_{1 \to 0}^{up}}$ \& ${F_{0 \to 1}^{up}}$ mapped to specific time step ${t}$ in a linear function. 

\subsection{Texture Mapping of {\methodname}}
\label{sec:textmap}
In this section, Texture Mapping is used to generate the intermediate proxy and map it with selected high-quality textures. 
{\methodname} first predicts the intermediate proxy and extracts texture blocks from two input frames. 
By flow guidance and calculated similarity
, {\methodname} indexes the high-quality mapping blocks. 

\subsubsection{Proxy Prediction and Texture Extraction}
Based on warped frames ${I_t^0}$ \& ${I_t^1}$, as displayed in Fig.~\ref{fig:ppl}, the proxy Prediction module generates the intermediate proxy ${\mathcal{Q}}$:
\begin{equation}
{\cal Q} = Convs(I_t^0,I_t^1),{\cal Q} \in {^{\frac{H}{2} \times \frac{W}{2} \times C}},
\label{eq:eq1}
\end{equation} 
where ${C}$ denotes the channel number of ${\mathcal{Q}}$. We use multilayer convolutions for the proxy prediction. 
Concurrently, the textures of two consecutive frames are extracted and divided into texture blocks with a specific size setting. 

To refine ${\mathcal{Q}}$ into the interpolated frame ${{\hat I}_t}$, {\methodname} uses the obtained proxy ${\mathcal{Q}}$ to index the textures blocks and map them into the proxy ${\mathcal{Q}}$.
Based on frames ${I_0}$, ${I_1}$, we obtain the texture ${\mathcal{T}_0}$, ${ \mathcal{T}_1}$:
\begin{equation}
{\mathcal{T}_0} = Convs({I_0}),{\mathcal{T}_1} = Convs({I_1}).
\label{eq:eq2}
\end{equation} 
We use convolutional layers as an extractor. 
According to the block size ${s}$, we respectively divide the texture ${\mathcal{T}_0}$ and ${ \mathcal{T}_1}$ into many texture blocks ${\mathcal{B}_0^{x,y}}$ and ${\mathcal{B}_1^{x,y}}$, which overlaps with each other. For example, ${\mathcal{B}_0^{x,y}}$ can be expressed as:
\begin{equation}
\mathcal{B}_0^{x,y} = Split({\mathcal{T}_0}|x,y,s),\mathcal{B}_0^{x,y} \in {\mathbb{R}^{s \times s \times C}},
\label{eq:eq3}
\end{equation} 
where ${Split(\mathcal{T}|x,y,s)}$ denotes that the texture ${\mathcal{T}}$ is divided into ${s \times s}$ texture-blocks, with ${\mathcal{B}^{x,y}}$ representing the texture block located at position index ${(x, y)}$. 
In addition, the proxy ${\mathcal{Q}}$ is similarly divided into blocks for the purpose of texture mapping,  represented by ${\mathcal{B}_q^{x,y}}$: 
\begin{equation}
\mathcal{B}_q^{x,y} = Split({\mathcal{Q}}|x,y,s),\mathcal{B}_q^{x,y} \in {\mathbb{R}^{s \times s \times C}}.
\label{eq:eq3-1}
\end{equation} 

\subsubsection{Flow-guided Block-Texture Searching}
\label{sec:flowg}
By estimating the motion between the given frames ${I_0 }$ \& ${ I_1}$, bidirection flows ${F_{0 \to t}^{up}}$ \& ${F_{1 \to t}^{up}}$ can be obtained. 
Since these flows contain direct motion information, they facilitate texture indexing. 
Therefore, to obtain preliminary texture block matching, we use flows ${F_{0 \to t}^{up}}$ and ${F_{1 \to t}^{up}}$ for indexing.

Firstly, according to block size ${s \times s}$, we respectively downsample ${F_{0 \to t}^{up}}$ and ${F_{1 \to t}^{up}}$ to obtain ${F_{0 \to t}^{block}}$ and ${F_{1 \to t}^{block}}$, in `nearest' mode.
To avoid texture mixing that can result from the application of bilinear-based forward or backward warping,  we first convert the forward flows ${F_{0 \to t}^{block}, F_{1 \to t}^{block}}$ into backward flows ${F_{t \to 0}^{block}, F_{t \to 1}^{block}}$: 
\begin{equation}
F_{t \to 0}^{block} =  - 1 \times {\cal{W}}(F_{0 \to t}^{block},F_{0 \to t}^{block}).
\label{eq:eq5}
\end{equation} 
Then, we use the `$Grid\_Sample$'(Nearest mode) function in PyTorch~\cite{paszke2019pytorch} to index texture blocks and obtain indexed texture blocks ${{\cal B}_{0,t}}$, which can be expressed as:
\begin{equation}
{{\cal B}_{0,t}} = GridSample({{\cal B}_0},F_{t \to 0}^{block},mode=Nearest).
\label{eq:eq6}
\end{equation} 
Following Eq.~\eqref{eq:eq5} and Eq.~\eqref{eq:eq6}, we obtain ${{\cal B}_{0,t}}$ and ${{{\cal B}_{1,t}}}$. These are rearrangement of the texture blocks ${{\cal B}_{0}}$ and ${{{\cal B}_{1}}}$.

\begin{table*}[t]
  \centering
  \scalebox{0.78}[0.7]
  {
\begin{tabular}{cccccccccc||ccc}
\toprule
\multirow{2}[4]{*}{} & \multirow{2}[4]{*}{Venue} & \multicolumn{4}{c}{DAVIS(1080p)} & \multicolumn{4}{c||}{DAVIS(4K)} & Vimeo90K  & \multirow{2}[4]{*}{R.} & \multirow{2}[4]{*}{F.} \\
\cmidrule(r){3-6} \cmidrule(r){7-10} \cmidrule(r){11-11}  &   & PSNR${\uparrow}$ & SSIM${\uparrow}$ & LPIPS${\downarrow}$ & DISTS${\downarrow}$ & PSNR${\uparrow}$ & SSIM${\uparrow}$ & LPIPS${\downarrow}$ & DISTS${\downarrow}$ & PSNR${\uparrow}$ &   &  \\
\midrule
XVFI & ICCV21 & 25.219  & 0.794  & 0.170  & 0.069  & 24.799  & 0.799  & 0.182  & 0.071  & 35.070  & 69 & 85 \\
RIFE & ECCV22 & 25.907  & 0.803  & \textcolor[rgb]{ 1,  0,  0}{\textbf{0.134 }} & \textcolor[rgb]{ 1,  0,  0}{\textbf{0.052 }} & 25.691  & 0.809  & \textcolor[rgb]{ 0,  0,  1}{\underline{0.143}} & \textcolor[rgb]{ 1,  0,  0}{\textbf{0.055 }} & 34.189  & \textcolor[rgb]{ 1,  0,  0}{\textbf{29}} & \textcolor[rgb]{ 1,  0,  0}{\textbf{52}} \\
M2M & CVPR22 & 26.191  & 0.812  & 0.164  & 0.072  & 25.900  & 0.816  & 0.173  & 0.076  & 35.369  & 61 & \textcolor[rgb]{ 0,  0,  1}{\underline{73}} \\
AMT-G & CVPR23 & 25.950  & 0.813  & 0.180  & 0.079  & OOM & OOM & OOM & OOM & \textcolor[rgb]{ 1,  0,  0}{\textbf{36.352 }} & 120 & 638 \\
UPR-Large & CVPR23 & 26.582  & 0.820  & 0.156  & 0.068  & 26.405  & 0.825  & 0.161  & 0.069  & 36.115  & 54 & 251 \\
UPR-LLarge & CVPR23 & 26.655  & 0.821  & 0.156  & 0.069  & OOM & OOM & OOM & OOM & \textcolor[rgb]{ 0,  0,  1}{\underline{36.247}} & 65 & 445 \\
EMA-VFI  & CVPR23 & 26.465  & 0.816  & 0.169  & 0.073  & 26.351  & 0.822  & 0.173  & 0.075  & 35.917  & \textcolor[rgb]{ 0,  0,  1}{\underline{44}} & 109 \\
SGM-Local & CVPR24 & 26.008  & 0.808  & 0.159  & 0.067  & 24.710  & 0.784  & 0.196  & 0.084  & 34.286  & 104 & 188 \\
SGM-1/2Point & CVPR24 & \textcolor[rgb]{ 0,  0,  1}{\underline{26.927}} & \textcolor[rgb]{ 0,  0,  1}{\underline{0.826}} & 0.149  & 0.068  & \textcolor[rgb]{ 0,  0,  1}{\underline{26.798}} & \textcolor[rgb]{ 0,  0,  1}{\underline{0.831}} & 0.154  & 0.071  & 35.670  & 108 & 188 \\
VTinker-${\mathcal{L}_{cc}}$ & -- & \textcolor[rgb]{ 1,  0,  0}{\textbf{26.976 }} & \textcolor[rgb]{ 1,  0,  0}{\textbf{0.827 }} & \textcolor[rgb]{ 1,  0,  0}{\textbf{0.134 }} & \textcolor[rgb]{ 0,  0,  1}{\underline{0.059}} & \textcolor[rgb]{ 1,  0,  0}{\textbf{26.826 }} & \textcolor[rgb]{ 1,  0,  0}{\textbf{0.833 }} & \textcolor[rgb]{ 1,  0,  0}{\textbf{0.139 }} & \textcolor[rgb]{ 0,  0,  1}{\underline{0.063}} & 35.640  & 121 & 765 \\
\midrule
SoftSplat & CVPR20 & 25.385  & 0.784  & 0.147  & 0.054  & \textcolor[rgb]{ 0,  0,  1}{\underline{25.018}} & \textcolor[rgb]{ 0,  0,  1}{\underline{0.788}} & \textcolor[rgb]{ 0,  0,  1}{\underline{0.159}} & \textcolor[rgb]{ 0,  0,  1}{\underline{0.058}} & \textcolor[rgb]{ 0,  0,  1}{\underline{35.359}} & \textcolor[rgb]{ 0,  0,  1}{\underline{117}} & \textcolor[rgb]{ 0,  0,  1}{\underline{268}} \\
EDSC & PAMI21 & 24.547  & 0.768  & 0.205  & 0.082  & 24.195  & 0.771  & 0.215  & 0.087  & 34.843  & \textcolor[rgb]{ 1,  0,  0}{\textbf{55}} & \textcolor[rgb]{ 1,  0,  0}{\textbf{72}} \\
FILM & ECCV22 & 26.009  & 0.801  & 0.124  & 0.048  & 23.078  & 0.709  & 0.252  & 0.099  & \textcolor[rgb]{ 1,  0,  0}{\textbf{35.550 }} & 273 & -- \\
PerVFI & CVPR24 & \textcolor[rgb]{ 0,  0,  1}{\underline{26.230}} & 0.808  & \textcolor[rgb]{ 0,  0,  1}{\underline{0.114}} & \textcolor[rgb]{ 0,  0,  1}{\underline{0.042}} & OOM & OOM & OOM & OOM & 33.841  & 285 & 872 \\
VTinker & -- & \textcolor[rgb]{ 1,  0,  0}{\textbf{26.778 }} & \textcolor[rgb]{ 1,  0,  0}{\textbf{0.817 }} & \textcolor[rgb]{ 1,  0,  0}{\textbf{0.108 }} & \textcolor[rgb]{ 1,  0,  0}{\textbf{0.039 }} & \textcolor[rgb]{ 1,  0,  0}{\textbf{26.610 }} & \textcolor[rgb]{ 1,  0,  0}{\textbf{0.823 }} & \textcolor[rgb]{ 1,  0,  0}{\textbf{0.115 }} & \textcolor[rgb]{ 1,  0,  0}{\textbf{0.041 }} & 35.064  & 121 & 765 \\
\bottomrule
\bottomrule
\end{tabular}%
} 
    \caption{Quantitative comparison of state-of-the-art VFI methods on DAVIS ({1080P}), DAVIS (\textbf{4K}), and Vimeo90K (448 ${\times}$ 256). OOM means Out Of Memory. R. and F. donate \textbf{R}untime(ms) and \textbf{F}LOPs(G). \textcolor[rgb]{ 1,  0,  0}{\textbf{RED}}: best performance, \textcolor[rgb]{ 0,  0,  1}{\underline{BLUE}}: the second best. Methods listed above the line are trained without perception
loss, while those below are trained with perception loss.}
  \label{tab:maintab}%
\end{table*}%

\begin{table*}[ht]
  \centering
  \scalebox{0.75}[0.8]
  {
\begin{tabular}{ccccccccccccc}
\toprule
\multirow{3}[6]{*}{} & \multicolumn{8}{c}{SNU-FILM}  & \multicolumn{2}{c}{\multirow{2}[4]{*}{Xiph-2K}} & \multicolumn{2}{c}{\multirow{2}[4]{*}{Xiph-4K}} \\
\cmidrule{2-9}  & \multicolumn{2}{c}{Easy} & \multicolumn{2}{c}{Medium} & \multicolumn{2}{c}{Hard} & \multicolumn{2}{c}{Extreme} & \multicolumn{2}{c}{} & \multicolumn{2}{c}{} \\
\cmidrule{2-13}  & LPIPS${\downarrow}$ & DISTS${\downarrow}$ & LPIPS${\downarrow}$ & DISTS${\downarrow}$ & LPIPS${\downarrow}$ & DISTS${\downarrow}$ & LPIPS${\downarrow}$ & DISTS${\downarrow}$ & LPIPS${\downarrow}$ & DISTS${\downarrow}$ & LPIPS${\downarrow}$ & DISTS${\downarrow}$ \\
\midrule
EDSC & 0.020  & 0.022  & 0.036  & 0.032  & 0.077  & 0.050  & 0.150  & 0.076  & 0.086  & 0.046  & 0.190  & 0.078  \\
FILM & 0.014  & 0.013  & \textcolor[rgb]{ 0,  0,  1}{\underline{0.023 }} & \textcolor[rgb]{ 0,  0,  1}{\underline{0.018 }} & \textcolor[rgb]{ 0,  0,  1}{\underline{0.046 }} & \textcolor[rgb]{ 0,  0,  1}{\underline{0.026 }} & \textcolor[rgb]{ 0,  0,  1}{\underline{0.093 }} & 0.045  & \textcolor[rgb]{ 0,  0,  1}{\underline{0.033 }} & 0.023  & 0.513  & 0.194  \\
RIFE & 0.014  & 0.013  & 0.025  & 0.019  & 0.050  & 0.029  & 0.101  & 0.049  & 0.040  & 0.020  & \textcolor[rgb]{ 0,  0,  1}{\underline{0.084 }} & \textcolor[rgb]{ 0,  0,  1}{\underline{0.035 }} \\
AMT-G & 0.020  & 0.023  & 0.035  & 0.034  & 0.062  & 0.048  & 0.124  & 0.072  & 0.096  & 0.053  & OOM & OOM \\
UPR-Large & 0.019  & 0.021  & 0.035  & 0.033  & 0.063  & 0.048  & 0.114  & 0.067  & 0.099  & 0.055  & 0.226  & 0.101  \\
UPR-LLarge & 0.019  & 0.021  & 0.035  & 0.034  & 0.064  & 0.049  & 0.114  & 0.068  & 0.101  & 0.055  & OOM & OOM \\
LDMVFI & \textcolor[rgb]{ 1,  0,  0}{\textbf{0.013 }} & -- & 0.027  & -- & 0.068${^{\dagger}}$ & -- & 0.139${^{\dagger}}$ & -- & -- & -- & -- & -- \\
EMA-VFI  & 0.019  & 0.021  & 0.035  & 0.033  & 0.073  & 0.050  & 0.147  & 0.083  & 0.096  & 0.052  & 0.222  & 0.097  \\
SGM-Local & 0.023  & 0.024  & 0.036  & 0.033  & 0.068  & 0.048  & 0.138  & 0.078  & 0.097  & 0.052  & 0.217  & 0.094  \\
SGM-1/2point & 0.019  & 0.021  & 0.034  & 0.032  & 0.065  & 0.047  & 0.125  & 0.072  & 0.097  & 0.053  & 0.222  & 0.097  \\
PerVFI & 0.015  & \textcolor[rgb]{ 0,  0,  1}{\underline{0.012 }} & 0.026  & \textcolor[rgb]{ 0,  0,  1}{\underline{0.018 }} & 0.049  & 0.027  & 0.094  & \textcolor[rgb]{ 0,  0,  1}{\underline{0.044 }} & 0.038  & \textcolor[rgb]{ 0,  0,  1}{\underline{0.015 }} & OOM & OOM \\
VTinker & \textcolor[rgb]{ 1,  0,  0}{\textbf{0.013 }} & \textcolor[rgb]{ 1,  0,  0}{\textbf{0.011 }} & \textcolor[rgb]{ 1,  0,  0}{\textbf{0.022 }} & \textcolor[rgb]{ 1,  0,  0}{\textbf{0.016 }} & \textcolor[rgb]{ 1,  0,  0}{\textbf{0.044 }} & \textcolor[rgb]{ 1,  0,  0}{\textbf{0.025 }} & \textcolor[rgb]{ 1,  0,  0}{\textbf{0.088 }} & \textcolor[rgb]{ 1,  0,  0}{\textbf{0.040 }} & \textcolor[rgb]{ 1,  0,  0}{\textbf{0.031 }} & \textcolor[rgb]{ 1,  0,  0}{\textbf{0.013 }} & \textcolor[rgb]{ 1,  0,  0}{\textbf{0.066 }} & \textcolor[rgb]{ 1,  0,  0}{\textbf{0.025 }} \\
\bottomrule
\bottomrule
\end{tabular}%
}
\caption{Quantitative comparison of state-of-the-art VFI methods on \textbf{720P} dataset (SNU-FILM), \textbf{2K} dataset (Xiph-2K), and \textbf{4K} dataset (Xiph-4K). The scores of LDMVFI are taken from their paper and are denoted by symbol ${^{\dagger}}$.
  }
  \label{tab:1080p}%
\end{table*}%

\subsubsection{Local Matching for Block-Texture}
\label{sec:localmatch}
Due to the inaccuracy of the flow estimation, the proxy block ${\mathcal{B}_q^{x,y}}$ generally corresponds to neither ${\mathcal{B}_{0,t}^{x,y}}$ nor to ${\mathcal{B}_{1,t}^{x,y}}$, which would significantly affect the interpolated result. 
To achieve a better matching, we process the proxy ${\mathcal{B}_q}$ and texture maps ${\mathcal{B}_{0,t}}$ and ${\mathcal{B}_{1,t}}$ into low-resolution index-tensors ${\mathcal{K}_q, \mathcal{K}_{0,t}, \mathcal{K}_{1,t}}$ by multilayer convolutions:
\begin{equation}
{\cal K}_q^{x,y} = Convs({\cal B}_q^{x,y}),{{\cal K}_q^{x,y}} \in {\mathbb{R}^{\frac{s}{2} \times \frac{s}{2} \times {C^{'}}}},
\label{eq:eq7}
\end{equation}
where ${C^{'}}$ denotes the channel number of ${\mathcal{K}_q}$. The same operation is applied to obtain ${\mathcal{K}_{0,t}}$, ${\mathcal{K}_{1,t}}$. 
Inspired by QKV mechanism~\cite{vaswani2017attention}, we convert index-tensors ${{\cal K}_q^{x,y}}$, ${{\cal K}_{0,t}^{x,y}}$ and ${{\cal K}_{1,t}^{x,y}}$ to index-vectors ${Q^{x,y}}$, ${K_0^{x,y}}$ and ${K_1^{x,y}}$ by
\begin{equation}
\begin{array}{l}
{Q^{x,y}} = Mean(Norm({\cal K}_q^{x,y})),\\
K_0^{x,y} = Mean(Norm({\cal K}_{0,t}^{x,y})),\\
K_1^{x,y} = Mean(Norm({\cal K}_{1,t}^{x,y})),
\end{array}
\label{eq:eq8}
\end{equation} 
where ${Q^{x,y}, K_0^{x,y}, K_1^{x,y} \in {\mathbb{R}^{1 \times 1 \times {C^{'}}}}}$. 
Then, based on ${Q^{x,y}}$, we match the most relevant vector in ${N \times N}$ neighbors of both two index-vector groups ${K_0}$ and ${K_1}$ at the position ${(x,y)}$. ${{\mathbb{C}^{x,y,(N)}}}$ denotes a set of correlations with ${Q^{x,y}}$: 
\begin{equation}
{\mathbb{C}^{x,y,(N)}} = {Q^{x,y}} \times [\mathbb{K}_0^{x,y,(N)} \cap \mathbb{K}_1^{x,y,(N)}],
\label{eq:eq9}
\end{equation} 
\begin{small}
\begin{equation}
{\mathbb{C}^{x,y,(N)}} = \{C_e^{x + i,y + j}, - \frac{N}{2} \le i,j \le \frac{N}{2},e = \{ 0,1\} \},
\label{eq:eq11}
\end{equation}
\end{small} 
where ${C_e^{x + i,y + j}={Q^{x,y}} \times K_e^{x + i,y + j}}$, and ${\mathbb{K}_0^{x,y,(N)}}$ denotes a set of ${N \times N}$ neighbors of ${\mathcal{K}_0^{x,y}}$: 
\begin{equation}
{\mathbb{K}_0^{x,y,(N)} = \{ K_0^{x + i,y + j}, - \frac{N}{2} \le i,j \le \frac{N}{2}\} }.
\label{eq:eq10}
\end{equation} 
Then, we evaluate the maximum element ${C_e^{x+i, y+i}}$ in the set ${{\mathbb{C}^{x,y,(N)}}}$, and get the position index ${(x+i, y+j, e)}$ of this element. 
Note that this step ensures that the texture block comes from either ${I_0}$ or ${I_1}$. 
Thus, the texture block corresponding to the position ${(x,y)}$ of the proxy ${\mathcal{Q}}$ is ${\mathcal{B}_e^{x+i,y+j}}$, which will be mapped to ${\mathcal{Q}^{x,y}}$ to recover the detail of the interpolated result. 
Finally, following the steps above, all the texture block ${\mathcal{B}_q^{x,y}}$ are traversed.

\subsection{Reconstruction Module and Loss Function} 
\label{sec:reconloss}
As shown in Fig.~\ref{fig:ppl}, to ensure the quality of the texture block ${\mathcal{B}_0^{x,y}}$ and ${\mathcal{B}_1^{x,y}}$, we supervise not only the interpolated result ${{\hat I}_t}$ with Ground Truth ${I_t}$ but also the texture ${{\mathcal{T}_0}}$ and ${{\mathcal{T}_1}}$. 
The reconstruction module is a UNet-like network and aims to transform the latent space into the image space. By using weight-shared reconstruction module, we obtain the result ${{\hat I}_0}$ and ${{\hat I}_1}$ by restoring texture ${{\mathcal{T}_0}}$ and ${{\mathcal{T}_1}}$, and supervise them with the input frames ${I_0}$ and ${I_1}$. These frames do not require reconstruction during inference.

We train our model with the Style loss ${\mathcal{L}_S}$ proposed in FILM~\cite{reda2022film}:
\begin{equation}
{{\cal L}_S} = {w_l}{{\cal L}_1} + {w_{VGG}}{{\cal L}_{VGG}} + {w_{Gram}}{{\cal L}_{Gram}},
\label{eq:eq12}
\end{equation} 

Three Style losses ${\mathcal{L}_S^{t}}$, ${\mathcal{L}_S^{0}}$,  ${\mathcal{L}_S^{1}}$ are computed separately using ${{\hat I}_t}$, ${{\hat I}_0}$, ${{\hat I}_1}$ and ${{I}_t}$, ${{I}_0}$, ${{I}_1}$.  We combine these Style losses to obtain the final loss ${\mathcal{L}_S^{all}}$:
\begin{equation}
\mathcal{L}_S^{all} = w_t \times \mathcal{L}_S^{t} + w_0 \times \mathcal{L}_S^{0} + w_1 \times \mathcal{L}_S^{1}, 
\label{eq:eq15}
\end{equation} 
where the weights (${w_t}$, ${w_0}$, ${w_1}$) are set empirically.

\begin{figure*}[t]
    \centering
    \includegraphics[width=\linewidth, height=0.31\linewidth]{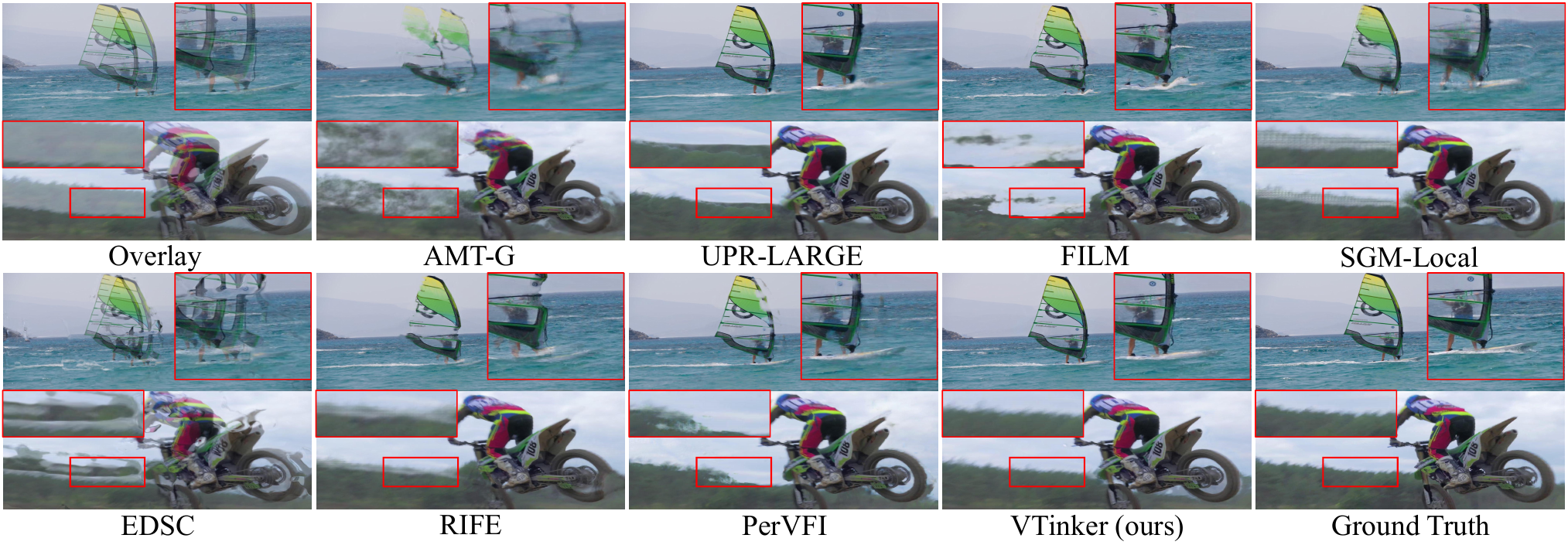}
    \caption{Qualitative comparisons among different methods on 2K resolution. All cases move more than 140 pixels. 
    Overlay is the average of two input frames. }
    \label{fig:viscom2k}
\end{figure*}

\section{Experiments}
\label{sec:exp}
\subsection{Experimental Settings.}
\noindent\textbf{Training Settings.}
{\methodname} is trained using the Style loss~\cite{reda2022film}, as defined in Eq.~\eqref{eq:eq15}, with the weights (${w_t}$, ${w_0}$, ${w_1}$) set to (0.8, 0.1, 0.1).
To ensure fair comparisons, using the commonly employed Charbonnier and Census Loss, we retrain VTinker, which is called VTinker-${\mathcal{L}_{cc}}$.
We utilize the commonly used Vimeo90K~\cite{xue2019video} Triplet as the training set and augment the images by randomly cropping 256 ${\times}$ 256 patches. 
AdamW optimizer~\cite{loshchilov2017decoupled} is used to train  {\methodname} for 0.2M iterations, with a batch size of 16 on 8 GPUs.
During training, we train the redesigned UPR-base model following the training setting of UPR-Net~\cite{jin2023unified}. 
Because of its lightweight design and efficiency, we use it as our motion estimator. 

\noindent \textbf{Compared Methods.}
We compare our {\methodname} with several state-of-the-art VFI algorithms, including methods trained without perception-based loss (XVFI~\cite{sim2021xvfi}, RIFE~\cite{huang2022real}, M2M~\cite{hu2022many}, AMT~\cite{li2023amt}, UPR~\cite{jin2023unified}, EMA-VFI~\cite{zhang2023extracting} and SGM~\cite{liu2024sparse}) and with percetion-based loss (SoftSplit~\cite{niklaus2020softmax}, EDSC~\cite{cheng2021multiple}, FILM~\cite{reda2022film}, LDMVFI~\cite{jain2024video}, and PerVFI~\cite{wu2024perception}). The publicly available implementations of those methods are used in comparisons. 

\noindent\textbf{Evaluation Metrics.}
We adopt the commonly used metrics for evaluation, including PSNR, SSIM, LPIPS~\cite{DBLP:conf/cvpr/ZhangIESW18}, DISTS~\cite{DBLP:journals/pami/DingMWS22}, and FloLPIPS~\cite{DBLP:journals/corr/abs-2207-08119}.
To address high LPIPS scores resulting from overfitting to VGG features, we use VGG19 for calculating loss while employing AlexNet for LPIPS computation.
Runtime and FLOPs are evaluated with a frame size of ${512 \times 512}$.

\noindent\textbf{Datasets.}
To evaluate the performance of the models, we employ commonly used VFI benchmarks: DAVIS (1080P, all full resolution cases are resized to 1080P)~\cite{DBLP:conf/cvpr/PerazziPMGGS16}, DAVIS(4K, includes large number of 4K cases with large motion), Vimeo90K~\cite{xue2019video} (${448\times256}$), SNU-FILM~\cite{choi2020channel} (${1280\times720}$, contains four subsets: easy, medium, hard and extreme, with increasing motion scales), Xiph~\cite{niklaus2020softmax}(2K, 4K), X-Test~\cite{sim2021xvfi}, UCF101~\cite{soomro2012ucf101} (${256\times256}$), and DAVIS (480P, ${640 \times 480}$).

\begin{figure*}[ht]
    \centering
    \includegraphics[width=\linewidth, height=0.3\linewidth]{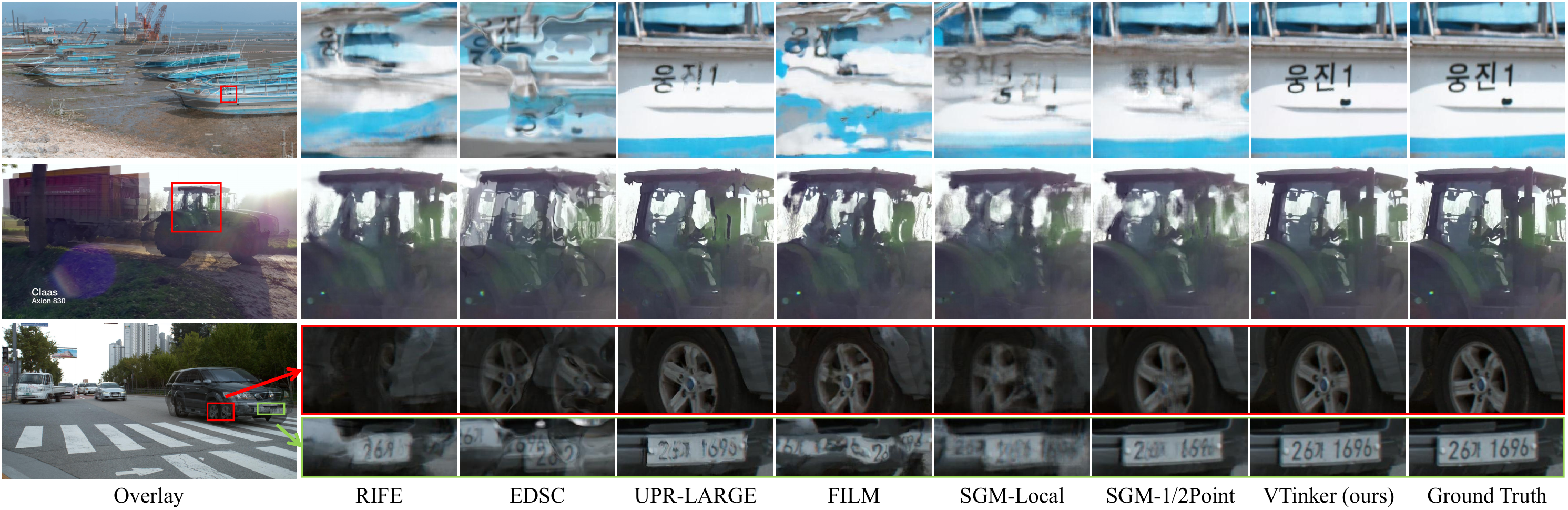}
    \caption{Qualitative comparisons among different methods on 4K. All cases move more than 200 pixels. Zoom in for details.}
    \label{fig:viscom4k}
\end{figure*}

\subsection{Quantitative Evaluation}
\label{sec:quaexp}
As shown in Tab.~\ref{tab:maintab}, according to whether perception-based loss is used in model training or not, we divide methods into two groups, and both versions of VTinker outperform others in PSNR, SSIM, and LPIPS for high-resolution VFI.
We also analyze {\methodname} in SNU-FILM and Xiph comparison with state-of-the-art methods in Tab.~\ref{tab:1080p}. The results show our proposed {\methodname} has the superior perceptual quality. 
In terms of the larger moving test scenarios (including SNU-FILM hard and extreme subsets, and DAVIS (1080P)), {\methodname} has more obvious advantages over the compared methods on SNU-FILM easy and medium subsets.
{\methodname} is trained only on Vimeo90K
, but performs better than the LDMVFI trained on high-resolution data, especially on SNU-FILM hard and extreme subsets. 
This also shows that our proposed {\methodname} has a strong generalizability.
For 2K resolution, Xiph-2K, the movement of this group is approximately 100 pixels. 
In contrast, for 4K resolution (Xiph-4K and DAVIS, the movement is around 200 pixels. 
Our proposed {\methodname} achieves an impressive performance in this comparison. 
For the 4K evaluation, the proposed {\methodname} obtains a 0.012 (approximately 34\%) improvement over the second-ranked method RIFE in terms of metric DISTS on the Xiph-4K dataset. 
The result highlights the efficacy of the proposed {\methodname} in the high-resolution VFI task.
More results, including comparisons on other benchmarks can be found in the supplementary material.

\begin{figure}[t]
    \centering
    \includegraphics[width=\linewidth, height=0.6\linewidth]{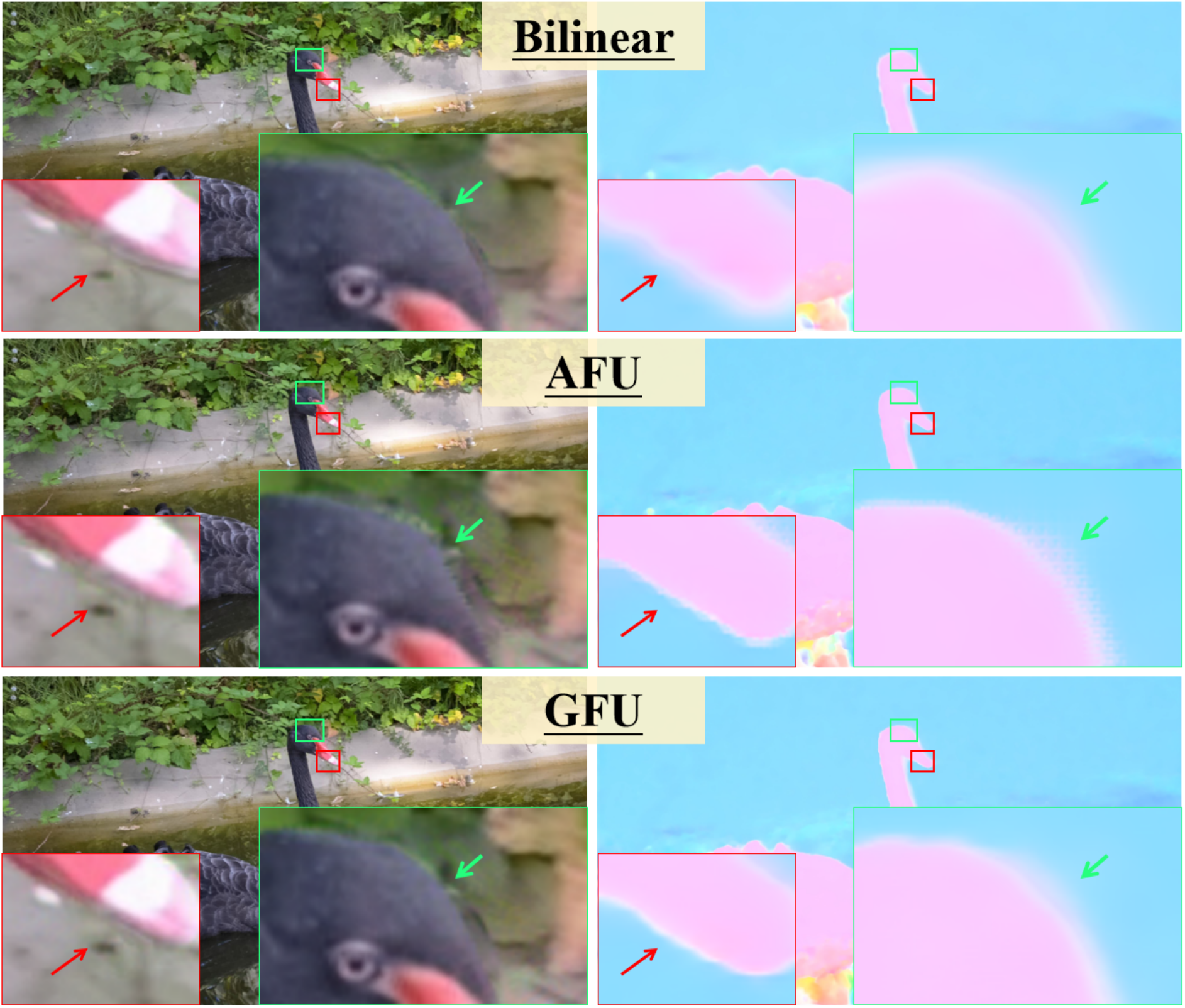}
    \caption{Comparison between various flow upsampling. Zoom in for mosaic-like and discontinuous details of the AFU. The left is the results, and the right shows the flow.} 
    \label{fig:xrsy}
\end{figure}

\subsection{Qualitative Evaluation}

In Fig.~\ref{fig:viscom2k}, all cases move over 140 pixels. 
Due to the misalignment of the estimated flow, the result of UPR-Large shows a wide range of ambiguities, such as the regions of sea and grass. 
{\methodname} uses high-quality texture mapping to recover detailed information in the intermediate proxy, which makes our result clearer.
In Fig.~\ref{fig:viscom4k}, motion distances in each case exceed 200 pixels. 
Facing large movement, some methods (such as UPR-Large, SGM-1/2point, and PerVFI) show blurring and artifacts, and some methods (such as RIFE and FILM) display distortion and ghosting in the results. 
It is worth noting that the results of the SGM (using AFU-like upsampling method) show discontinuities in the boundaries (`mosaic-like artifacts), which is due to the discontinuity in the upsampling flow processed by AFU (as displayed in Fig.~\ref{fig:xrsy}). 
Notably, the proposed {\methodname} exhibits impressive performance in the details, such as text. 

\begin{table}[ht]
  \centering
  \scalebox{0.855}
  {
\begin{tabular}{cccc}
\toprule
T.M. & Upsampling & Xiph-4K & Xiph-2K \\
\midrule
${\times}$ & Bilinear & 0.079/0.036 & 0.040/0.019 \\
${\times}$ & AFU & 0.081/0.031 & 0.039/0.017 \\
${\times}$ & GFU & 0.082/0.033 & 0.038/0.016 \\
\midrule
${\surd}$ & Bilinear & 0.080/0.034 & 0.037/0.018 \\
${\surd}$ & AFU & 0.078/0.031 & 0.037/0.017 \\
${\surd}$ & GFU & \textcolor[rgb]{ 1,  0,  0}{\textbf{0.066/0.025}} & \textcolor[rgb]{ 1,  0,  0}{\textbf{0.031/0.013}} \\
\Xhline{1.5px}
  & \multicolumn{2}{c}{w/o GFU ${\Rightarrow}$ GFU} & VS\\
\midrule
PSNR(Edge)${\uparrow}$ & \multicolumn{2}{c}{22.640 ${\Rightarrow}$ 25.070} & \textcolor[rgb]{ 1,  0,  0}{\boldmath{}\textbf{10.73\%${\uparrow}$}\unboldmath{}} \\
SSIM(Edge)${\uparrow}$ & \multicolumn{2}{c}{0.7575 ${\Rightarrow}$ 0.8143} & \textcolor[rgb]{ 1,  0,  0}{\boldmath{}\textbf{7.50\%${\uparrow}$}\unboldmath{}} \\
IoU(Edge)${\uparrow}$ & \multicolumn{2}{c}{0.2343 ${\Rightarrow}$ 0.3926} & \textcolor[rgb]{ 1,  0,  0}{\boldmath{}\textbf{67.56\%${\uparrow}$}\unboldmath{}} \\
\bottomrule
\bottomrule
\end{tabular}%
    }
\caption{Results of ablation experiments. Above the bold line: we show the metrics ``LPIPS ${\downarrow}$ / DISTS ${\downarrow}$'' on the dataset Xiph-2K and Xiph-4K~\cite{niklaus2020softmax}. T.M. denotes Texture Mapping. Below the bold line: comparison between w/o and w/ GFU on edge-related metrics. \textcolor[rgb]{ 1,  0,  0}{\textbf{RED}}: best performance}
  \label{tab:xrsy}%
\end{table}%

\subsection{Ablation Experiments}

\noindent \textbf{Guided Flow Upsampling.}
In the GFU module, we introduce the input frame information as a reference to make the boundary of the upsampled flow more consistent with the input frame. 
To verify the validity of the GFU module, we replace the GFU module with AFU~\cite{Luo2021ASFlowUO} and bilinear upsampling to train our model.
According to Tab.~\ref{tab:xrsy}, the use of GFU leads to better performance than AFU and bilinear upsampling. 
As shown in Fig.~\ref{fig:xrsy}, the AFU estimates a ${3 \times 3}$ sampling kernel for each upsampled target pixel, which leads to discontinuities in the boundaries of both upsampling flow and result.
Due to the model with AFU being trained on Vimeo90K, which is low-resolution, it does not generalize well to high resolution. 
In contrast, although {\methodname} is also trained on Vimeo90K, GFU achieves better high-resolution upsampling results, demonstrating its strong generalization capability.
Our GFU module is modified based on the bilinear upsampling, and we introduce the input frame information to refine the upsampled boundary. 
When GFU and Texture Mapping are used together, the overall gain from GFU is minimal when evaluated with whole-image metrics. However, its advantages become more apparent in edge-region evaluations and visualizations.
To assess performance in edge regions, we first apply the Canny operator to extract an edge mask from the image. We then compute PSNR, SSIM, and IoU metrics within this masked region.
In Tab.~\ref{tab:xrsy}, it shows that GFU improves the quality of the edge region of the inference result significantly.
As displayed in Fig.~\ref{fig:xrsy}, compared to the retrained model that uses bilinear upsampling, GFU produces sharper flow boundaries, resulting in clearer and more detailed outputs.

\noindent \textbf{Texture Mapping.}
After processing with texture mapping, the region with ghosting and blurring is mapped by a high-quality block-texture, which improves the clarity of the interpolated results. 
As displayed in Tab.~\ref{tab:xrsy}, the proposed {\methodname} trained without texture mapping would lead to considerable degradation of the model's performance. 
In particular, texture mapping can improve the quality of detail in the interpolated results for high-resolution interpolation. 

\section{Conclusion}
\label{sec:conc}
We proposed {\methodname}, which consists of two core components: GFU and Texture Mapping. Extensive experiments have demonstrated that {\methodname} outperforms the state-of-the-art VFI methods and produces high-quality intermediate frames, especially in high-resolution video.

\section{Acknowledgments}
This work was supported in part by the National Natural Science Foundation of China (62306153, 62225604), the Natural Science Foundation of Tianjin, China (24JCJQJC00020), the Young Elite Scientists Sponsorship Program by CAST (YESS20240686), the Fundamental Research Funds for the Central Universities (Nankai University, 070-63243143),  and Shenzhen Science and Technology Program (JCYJ20240813114237048). The computational devices are partly supported by the Supercomputing Center of Nankai University (NKSC). This work was also supported by the OPPO Research Fund.
\bibliography{arxiv}

\clearpage
\appendix

\section{Quantitative Comparison}
\label{sec:moreexp}
In this section, we give more experimental results, including quantitative comparisons on LRSL and X-Test~\cite{sim2021xvfi}. 
LRSL compares various state-of-the-art methods at low resolution. As shown in Tab.~\ref{tab:nomal}, following the comparison setup of PerVFI~\cite{wu2024perception}, we use PSNR, LPIPS~\cite{DBLP:conf/cvpr/ZhangIESW18}, DISTS~\cite{DBLP:journals/pami/DingMWS22}, and FloLPIPS~\cite{DBLP:journals/corr/abs-2207-08119} to compare the proposed {\methodname} with state-of-the-art methods. 
As displayed in Tab.~\ref{tab:x24K}, we also report LPIPS~\cite{DBLP:conf/cvpr/ZhangIESW18} and DISTS~\cite{DBLP:journals/pami/DingMWS22} evaluated on X-Test(2K)~\cite{sim2021xvfi} and X-Test(4K)~\cite{sim2021xvfi}.  We detail the experimental results as follows.

\noindent \textbf{Evalution on LRSL:} 
For LRSL, we evaluate {\methodname} at lower resolution. As shown in Tab.~\ref{tab:nomal}, we conduct comparisons for the LRSL using three datasets: Vimeo90K~\cite{xue2019video}(448 ${\times}$ 256), UCF101~\cite{soomro2012ucf101}(256 ${\times}$ 256), and DAVIS(480P, 854 ${\times}$ 480)~\cite{DBLP:conf/cvpr/PerazziPMGGS16}. These datasets represent scenarios with relatively low resolution and limited motion, resulting in minimal performance differences among the evaluated methods in these conditions. Despite this, ours {\methodname} still achieves good performance in the comparison. 

\begin{table*}[t]
  \centering
  \caption{Quantitative comparison of VFI state-of-the-art methods on Vimeo90K~\cite{xue2019video}, DAVIS-2017~\cite{DBLP:conf/cvpr/PerazziPMGGS16} (480P), and UCF101~\cite{soomro2012ucf101}, which tests methods' performance at \textbf{LRSL} resolutions. The scores of LDMVFI~\cite{jain2024video} are taken from their paper and are denoted by ${^{\dagger}}$. \textcolor[rgb]{ 1,  0,  0}{\textbf{RED}}: best performance, \textcolor[rgb]{ 0,  0,  1}{\underline{BLUE}}: second best performance. Although the proposed {\methodname} is specifically designed for high-resolution VFI, it achieves comparable or even superior performance compared to methods better suited for lower-resolution VFI, particularly in terms of perceptual quality metrics.}
  \scalebox{0.89}
  { 
\begin{tabular}{ccccccccccc}
\toprule
\multirow{2}[4]{*}{} & \multicolumn{3}{c}{Vimeo90K} & \multicolumn{3}{c}{UCF101} & \multicolumn{4}{c}{DAVIS(480P)} \\
\cmidrule{2-11}  & PSNR${\uparrow}$ & LPIPS${\downarrow}$ & DISTS${\downarrow}$ & PSNR${\uparrow}$ & LPIPS${\downarrow}$ & DISTS${\downarrow}$ & PSNR${\uparrow}$ & LPIPS${\downarrow}$ & DISTS${\downarrow}$ & FloLPIPS${\downarrow}$ \\
\midrule
EDSC & 34.843  & 0.027  & 0.042  & 35.133  & 0.030  & 0.044  & 26.359  & 0.134  & 0.070  & 0.094  \\
FILM & 35.550  & \textcolor[rgb]{ 1,  0,  0}{\textbf{0.014 }} & \textcolor[rgb]{ .282,  .455,  .796}{\textbf{0.023 }} & 34.862  & \textcolor[rgb]{ 1,  0,  0}{\textbf{0.022 }} & \textcolor[rgb]{ .282,  .455,  .796}{\textbf{0.035 }} & 27.048  & \textcolor[rgb]{ 1,  0,  0}{\textbf{0.074 }} & \textcolor[rgb]{ .282,  .455,  .796}{\textbf{0.041 }} & \textcolor[rgb]{ 1,  0,  0}{\textbf{0.054 }} \\
RIFE & 34.189  & 0.020  & 0.028  & 34.760  & 0.024  & 0.037  & 26.802  & 0.087  & 0.047  & 0.064  \\
AMT-G & \textcolor[rgb]{ 1,  0,  0}{\textbf{36.352 }} & 0.020  & 0.033  & 35.128  & 0.032  & 0.047  & \textcolor[rgb]{ 1,  0,  0}{\textbf{27.676 }} & 0.098  & 0.059  & 0.071  \\
UPR-Large & 36.115  & 0.021  & 0.035  & \textcolor[rgb]{ .282,  .455,  .796}{\textbf{35.154 }} & 0.032  & 0.047  & 27.349  & 0.156  & 0.068  & 0.085  \\
UPR-LLarge & \textcolor[rgb]{ .282,  .455,  .796}{\textbf{36.247 }} & 0.021  & 0.035  & \textcolor[rgb]{ 1,  0,  0}{\textbf{35.161 }} & 0.032  & 0.047  & 27.436  & 0.119  & 0.068  & 0.084  \\
LDMVFI & - & - & - & 32.160  & 0.026  & - & 25.073  & 0.125  & - & 0.172  \\
EMA-VFI  & 35.917  & 0.022  & 0.035  & 35.084  & 0.031  & 0.045  & 27.193  & 0.127  & 0.071  & 0.088  \\
SGM-Local & 34.286  & 0.025  & 0.038  & 32.561  & 0.034  & 0.048  & 26.916  & 0.117  & 0.066  & 0.081  \\
SGM-1/2point & 35.670  & 0.022  & 0.036  & 35.083  & 0.031  & 0.045  & \textcolor[rgb]{ .282,  .455,  .796}{\textbf{27.616 }} & 0.105  & 0.064  & 0.075  \\
PerVFI & 33.841  & 0.019  & 0.025  & 33.293  & 0.028  & 0.040  & 26.827  & \textcolor[rgb]{ .282,  .455,  .796}{\textbf{0.077 }} & 0.042  & 0.058  \\
VTinker & 35.064  & \textcolor[rgb]{ .282,  .455,  .796}{\textbf{0.015 }} & \textcolor[rgb]{ 1,  0,  0}{\textbf{0.022 }} & 34.714  & \textcolor[rgb]{ .282,  .455,  .796}{\textbf{0.023 }} & \textcolor[rgb]{ 1,  0,  0}{\textbf{0.035 }} & 27.208  & \textcolor[rgb]{ 1,  0,  0}{\textbf{0.074 }} & \textcolor[rgb]{ 1,  0,  0}{\textbf{0.038 }} & \textcolor[rgb]{ .282,  .455,  .796}{\textbf{0.055 }} \\
\bottomrule
\bottomrule
\end{tabular}%

}
  \label{tab:nomal}%
\end{table*}%

AMT-G~\cite{li2023amt} achieves the highest PSNR in the test scenarios of Vimeo90K~\cite{xue2019video} and DAVIS(480P)~\cite{DBLP:conf/cvpr/PerazziPMGGS16}, showing its notable capability to maintain pixel-level consistency during video frame interpolation. Similarly, the UPR-LLarge~\cite{jin2023unified} demonstrates superior performance on the UCF101~\cite{soomro2012ucf101} dataset, achieving the highest PSNR, which further highlights its robustness in addressing low-resolution interpolation task effectively. However, in terms of perceptual quality, since the training process is supervised using the style loss, the FILM~\cite{reda2022film} achieves the highest LPIPS scores across all three low-resolution datasets. However, as shown in Tab.~\ref{tab:maintab} of the main paper and Tab.~\ref{tab:x24K}, its performance is degraded in the face of high-resolution frame interpolation due to possessing a poor generalization to high resolutions (such as 4K). 
Although the proposed {\methodname} is specifically designed for high-resolution VFI, it achieves comparable or even superior performance compared to methods better suited for lower-resolution VFI, especially in terms of perceptual quality metrics.

\noindent \textbf{Evalution on X-Test:}
The X-Test~\cite{sim2021xvfi} dataset is derived from X4K1000FPS~\cite{sim2021xvfi}, a benchmark designed specifically for 4K-resolution video, renowned for its exceptional clarity and video quality. To comprehensively evaluate the performance of different interpolation methods, we conduct tests on two versions of the X-Test~\cite{sim2021xvfi} dataset: a downscaled 2K version (resized from the original X-Test) and the original 4K version. This setup allows us to assess the adaptability and effectiveness of each method across varying resolutions. 

To ensure a fair and consistent comparison, we use a recursive interpolation approach to generate continuous intermediate frames for all methods, thereby ensuring that all methods are evaluated under identical conditions. For example, to obtain the frame at the 0.25 time-step, we first interpolate the frame at the 0.5 time-step and then perform interpolation between the frames at 0 and 0.5 time-step. We assess performance using perceptual quality metrics, including LPIPS~\cite{DBLP:conf/cvpr/ZhangIESW18} and DISTS~\cite{DBLP:journals/pami/DingMWS22}. As shown in Tab.~\ref{tab:x24K}, the proposed method {\methodname} consistently outperforms state-of-the-art methods on both the 2K and 4K versions of the X-Test~\cite{sim2021xvfi} dataset. The superior performance of {\methodname}, especially at higher resolutions, highlights our method's robust capability to preserve visual quality and fine details in high-resolution video interpolation tasks. These results further emphasize the advantages of our approach when dealing with high-resolution video content. While many methods struggle to maintain quality as resolution increases, our method demonstrates a clear edge, making it highly suitable for modern applications that demand precise and high-quality frame interpolation in 4K and beyond.

\begin{table}[t]
  \centering
  \caption{Quantitative comparison of VFI state-of-the-art methods on X-Test~\cite{sim2021xvfi}. OOM means Out Of Memory. }
\scalebox{1}{
\begin{tabular}{ccccc}
\toprule
  & \multicolumn{2}{c}{X-Test (2K)} & \multicolumn{2}{c}{X-Test (4K)} \\
\cmidrule{2-5}  & LPIPS${\downarrow}$ & DISTS${\downarrow}$ & LPIPS${\downarrow}$ & DISTS${\downarrow}$ \\
\midrule
EDSC & 0.1867  & 0.1010  & 0.2032  & 0.1192  \\
FILM & 0.1357  & 0.0489  & 0.5806  & 0.2445  \\
RIFE & 0.1347  & 0.0512  & \textcolor[rgb]{ .282,  .455,  .796}{0.1676 } & \textcolor[rgb]{ .282,  .455,  .796}{0.0822 } \\
AMT-G & 0.1817  & 0.9590  & OOM & OOM \\
UPR-Large & 0.1397  & 0.0569  & 0.1743  & 0.0835  \\
UPR-LLarge & 0.1393  & 0.0570  & OOM & OOM \\
EMA-VFI  & 0.1592  & 0.0807  & 0.1861  & 0.1053  \\
SGM-Local & 0.1557  & 0.0761  & 0.1941  & 0.1066  \\
SGM-1/2point & 0.1483  & 0.0690  & 0.1855  & 0.1019  \\
PerVFI & \textcolor[rgb]{ .282,  .455,  .796}{0.1307 } & \textcolor[rgb]{ .282,  .455,  .796}{0.0398 } & OOM & OOM \\
VTinker & \textcolor[rgb]{ 1,  0,  0}{\textbf{0.1285 }} & \textcolor[rgb]{ 1,  0,  0}{\textbf{0.0383 }} & \textcolor[rgb]{ 1,  0,  0}{\textbf{0.1593 }} & \textcolor[rgb]{ 1,  0,  0}{\textbf{0.0510 }} \\
\bottomrule
\bottomrule
\end{tabular}%
}
  \label{tab:x24K}%
\end{table}%

\section{More Visual Results}
In this section, we provide additional visualization results.
We test our methods against other state-of-the-art methods like EDSC~\cite{cheng2021multiple}, FILM~\cite{reda2022film}, RIFE~\cite{huang2022real}, AMT-G~\cite{li2023amt}, UPR-LARGE~\cite{jin2023unified}, SGM-1/2Point~\cite{liu2024sparse}, and PerVFI~\cite{wu2024perception} on the DAVIS~\cite{DBLP:conf/cvpr/PerazziPMGGS16}(with 4K resolution) and X-Test(4K)~\cite{sim2021xvfi} for qualitative evaluation. 
The specific visualization results can be found in Figs.~\ref{fig:supvis0}-\ref{fig:supvis4}. 
It is observed that the proposed {\methodname} outperforms other methods in the context of high-resolution VFI, resulting in the generation of more discernible outcomes. 

\section{Discussion on GFU Module}
As mentioned in the section (Related Work) of the main paper, when VFI models face high-resolution video, almost all methods first estimate the flow at the low resolution and then warp the original high-resolution by the upsampled flows.
However, when the upsampling methods are bilinear(used by most methods), the edge of the upsampled flow is misaligned with the image's boundaries, which leads to blurred or incorrect edges in the results.
The design of GFU is inspired by UPFlow~\cite{luo2021upflow}, which uses guidance information to help make backward flow upsampling more accurate for the flow estimation task.
However, the backward flow's edges are aligned with the predicted result in VFI task, such as flow ${F_{t \to 0}}$ is aligned with the predicted frame ${I_{t}}$(we cannot get it), but either the input frame ${I_{0}}$ or ${I_{1}}$.
Thus, when faced with \underline{video frame interpolation tasks}, its guidance information is hard to be effective.
To make more efficient use of this guidance information, we use a forward flow-based framework, whose edges of the flow are strictly aligned with the input frame, such as between the forward flow ${F_{0 \to 1}}$ and the frame ${I_{0}}$.
Utilizing the nature of the forward flow, we design GFU and achieve the extraction of guidance information using only a few convolutional layers. 
Noticeably, UPFlow guides \underline{backward flow} upsampling for \underline{flow estimation task}, but GFU upsamples \underline{forward flow} using guidance information for \underline{video frame interpolation task}, which was not used before.
As shown in Table 6 and Figure 3 of the main paper, this is a solid illustration of the better generalization and efficiency of GFU.

\begin{figure}[t]
    \centering
    \includegraphics[width=\linewidth]{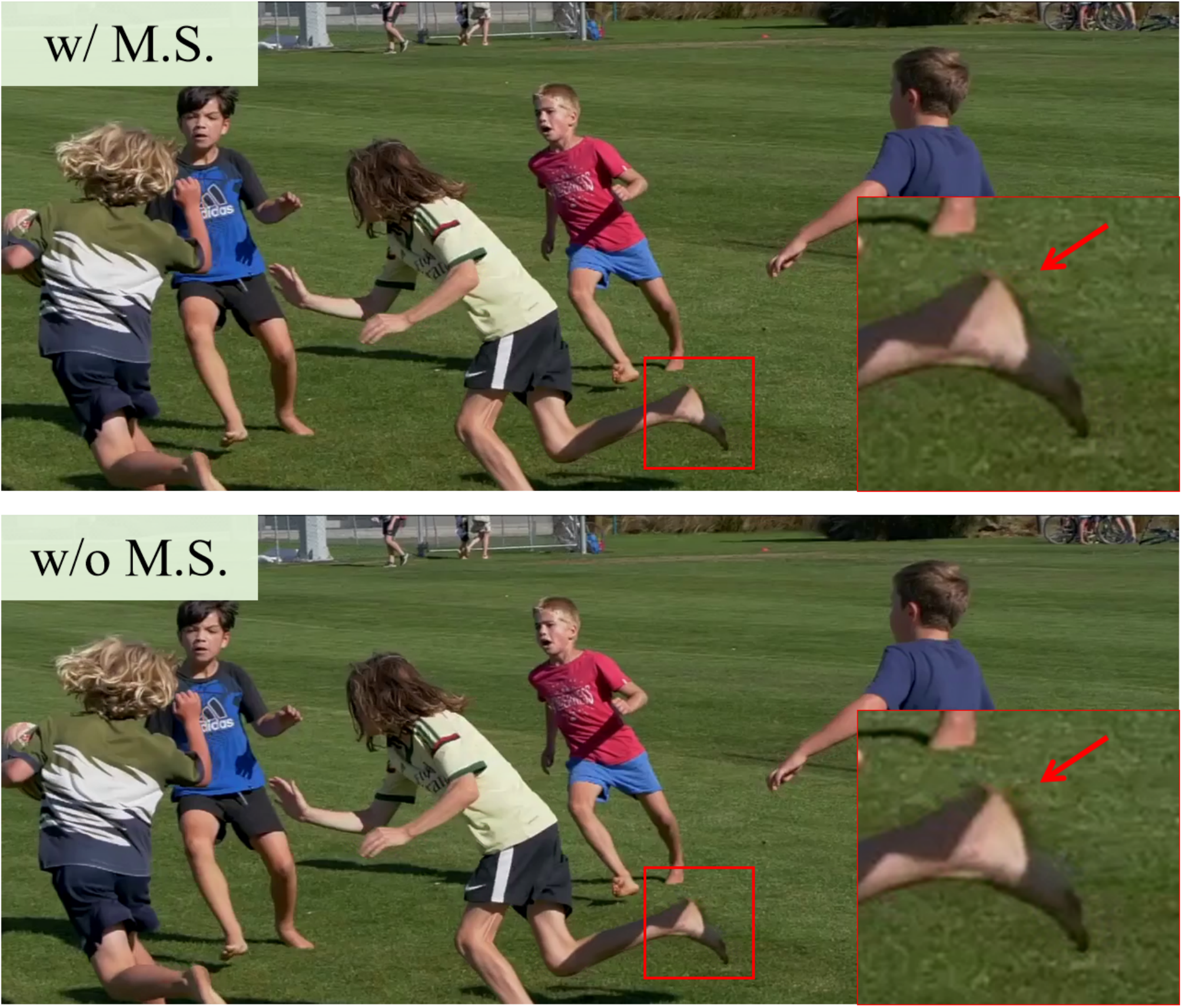}
    \caption{Qualitative comparison with and without Multi-supervisory. M.S. means Multi-supervisory. Zoom in for details. }
    \label{fig:xrsy_supms}
\end{figure}

\begin{table*}[ht]
  \centering
  \caption{Ablation experiment on loss function. \textbf{Bold}: best performance. P.L. means Perception-based Loss. M.S. means Multi-supervisory. }
    \begin{tabular}{cccccccc}
    \toprule
      &   & \multicolumn{2}{c}{DAVIS-480P} & \multicolumn{2}{c}{Xiph-2K} & \multicolumn{2}{c}{DAVIS(Full Res.)} \\
\cmidrule{3-8}    P.L. & M.S. & LPIPS${\downarrow}$ & DISTS${\downarrow}$ & LPIPS${\downarrow}$ & DISTS${\downarrow}$ & LPIPS${\downarrow}$ & DISTS${\downarrow}$ \\
    \midrule
    ${\times}$ & ${\surd}$ & 0.083  & 0.046  & 0.051  & 0.029  & 0.124  & 0.051  \\
    ${\surd}$ & ${\times}$ & \textbf{0.074 } & 0.038  & \textbf{0.030 } & 0.012  & \textbf{0.115 } & \textbf{0.041 } \\
    ${\surd}$ & ${\surd}$ & \textbf{0.074 } & \textbf{0.037 } & \textbf{0.030 } & \textbf{0.011 } & \textbf{0.115 } & \textbf{0.041 } \\
    \bottomrule
    \bottomrule
    \end{tabular}%
  \label{tab:abladd}%
\end{table*}%

\begin{figure*}[t]
    \centering
    \includegraphics[width=\linewidth]{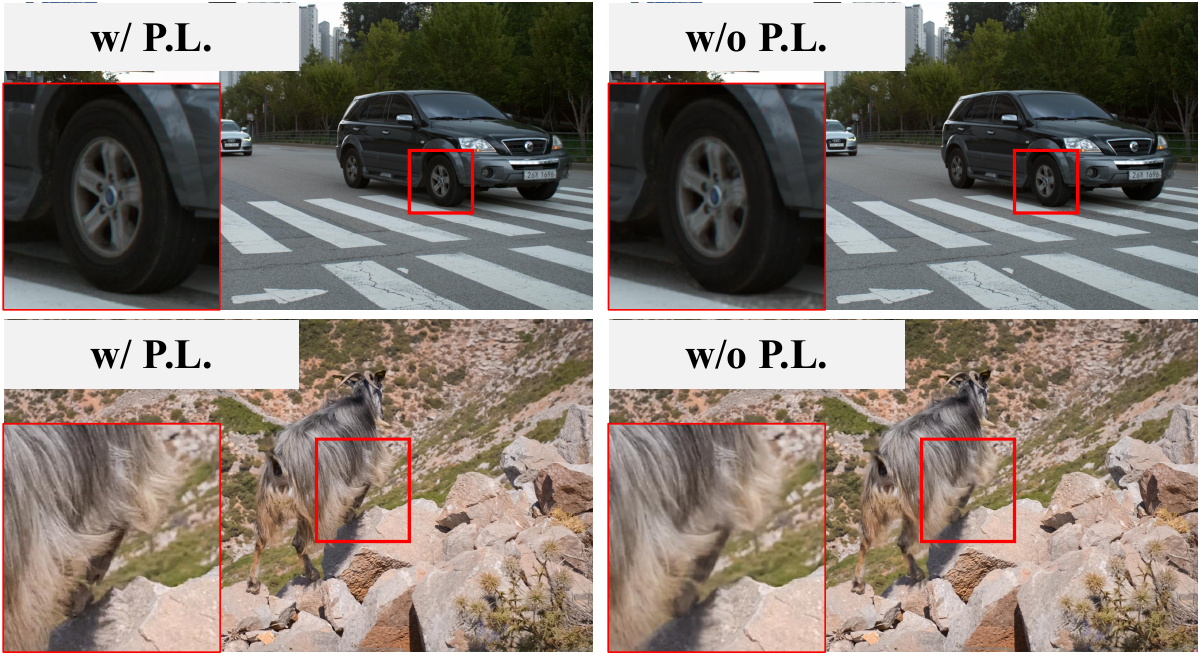}
    \caption{Qualitative comparison with and without Perception-based Loss. P.L. means Perception-based Loss. Zoom in for details. }
    \label{fig:xrsy_suppl}
    \vspace{-0.6cm}
\end{figure*}

\section{Ablation Experiment on Loss Function}
In the main paper, we describe the used loss function, Style loss ${\mathcal{L}_S}$ as:
\begin{equation}
{{\cal L}_S} = {w_l}{{\cal L}_1} + {w_{VGG}}{{\cal L}_{VGG}} + {w_{Gram}}{{\cal L}_{Gram}}.
\label{eq:eq31}
\end{equation} 
where (${w_l}$, ${w_{VGG}}$, ${w_{Gram}}$) means the weights of (${{\cal L}_1}$, ${{\cal L}_{VGG}}$, ${{\cal L}_{Gram}}$) and are set empirically
To ensure the quality of the texture block ${\mathcal{B}_0^{x,y}}$ and ${\mathcal{B}_1^{x,y}}$, we supervise not only the interpolated result ${{\hat I}_t}$ with Ground Truth ${I_t}$ but also the texture ${{\mathcal{T}_0}}$ and ${{\mathcal{T}_1}}$. Three Style losses ${\mathcal{L}_S^{t}}$, ${\mathcal{L}_S^{0}}$,  ${\mathcal{L}_S^{1}}$ are computed separately using ${{\hat I}_t}$, ${{\hat I}_0}$, ${{\hat I}_1}$ and ${{I}_t}$, ${{I}_0}$, ${{I}_1}$.  We combine these Style losses to obtain the final loss ${\mathcal{L}_S^{all}}$:
\begin{equation}
\mathcal{L}_S^{all} = 0.8 \times \mathcal{L}_S^{t} + 0.1 \times \mathcal{L}_S^{0} + 0.1 \times \mathcal{L}_S^{1}.
\label{eq:eq32}
\end{equation} 
In this section, we focus on two discussions about Eq.~\eqref{eq:eq31} and Eq.~\eqref{eq:eq32}. The details are presented below.

\noindent \textbf{Perception-based Loss: } As displayed in Eq.~\eqref{eq:eq31}, it is worth noting that the Style loss ${\mathcal{L}_S}$ includes perception-based components such as ${{\cal L}_{VGG}}$ and ${{\cal L}_{Gram}}$. Without any perception-based components, we redesign the loss function: 
\begin{equation}
{{\cal L}_S^{'}} = {{\cal L}_1},
\label{eq:eq33}
\end{equation} 
We retrain the proposed {\methodname} with the redesigned loss ${{\cal L}_S^{'}}$ (Eq.~\eqref{eq:eq33}), the experimental results can be found in Tab.~\ref{tab:abladd} and Fig.~\ref{fig:xrsy_suppl}. 
As displayed in Tab.~\ref{tab:abladd}, the performance of the model trained without perception-based loss components 
performs poorly on the perception metrics (LPIPS and DISTS). Meanwhile, as shown in Fig.~\ref{fig:xrsy_suppl}, visualization results of the redesigned model without perception-based loss components appear blurred in detail, which illustrates that the use of the perception-based loss components can obtain clear frame interpolation results.  

\noindent \textbf{Multi-supervisory Loss: } As shown in Eq.~\eqref{eq:eq32}, we retrain the proposed {\methodname} with another redesigned loss:
\begin{equation}
\mathcal{L}_S^{all'} = \mathcal{L}_S^{t}.
\label{eq:eq34}
\end{equation} 
In contrast to ${\mathcal{L}_S^{all}}$, ${\mathcal{L}_S^{all'}}$ (Eq.~\eqref{eq:eq34}) removes the supervision of between ${{\hat I}_0}$, ${{\hat I}_1}$ and ${{I}_0}$, ${{I}_1}$. We also retrain the proposed model {\methodname} with redesigned loss ${\mathcal{L}_S^{all'}}$. 
As shown in Tab.~\ref{tab:abladd}, we find that multi-supervision of the model achieves the best performance results in all scenarios, but there is a slight quantitative improvement compared to using single-supervised loss ${\mathcal{L}_S^{all'}}$ (Eq.~\eqref{eq:eq34}). As displayed in Fig.~\ref{fig:xrsy_supms}, the implementation of multi-supervision can enhance the clarity of the frame interpolation result, which improves the performance of the {\methodname}.

\begin{figure*}[h]
    \centering
    \includegraphics[width=\linewidth, height=1.25\linewidth]{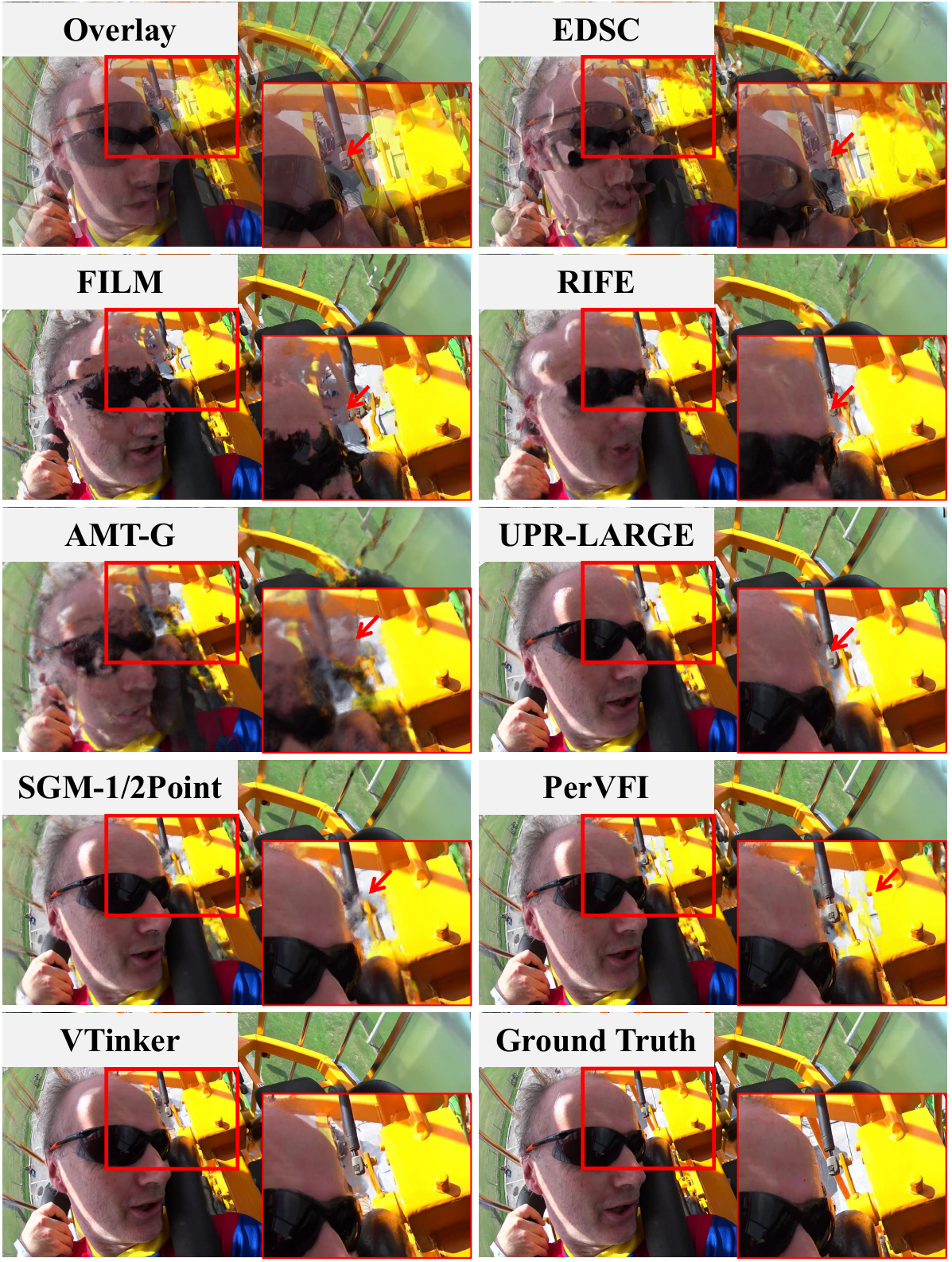}
    \caption{Qualitative comparisons among different methods on 2K resolution. Overlay is the average of two input frames. }
    \label{fig:supvis0}
\end{figure*}

\begin{figure*}[h]
    \centering
    \includegraphics[width=\linewidth, height=1.25\linewidth]{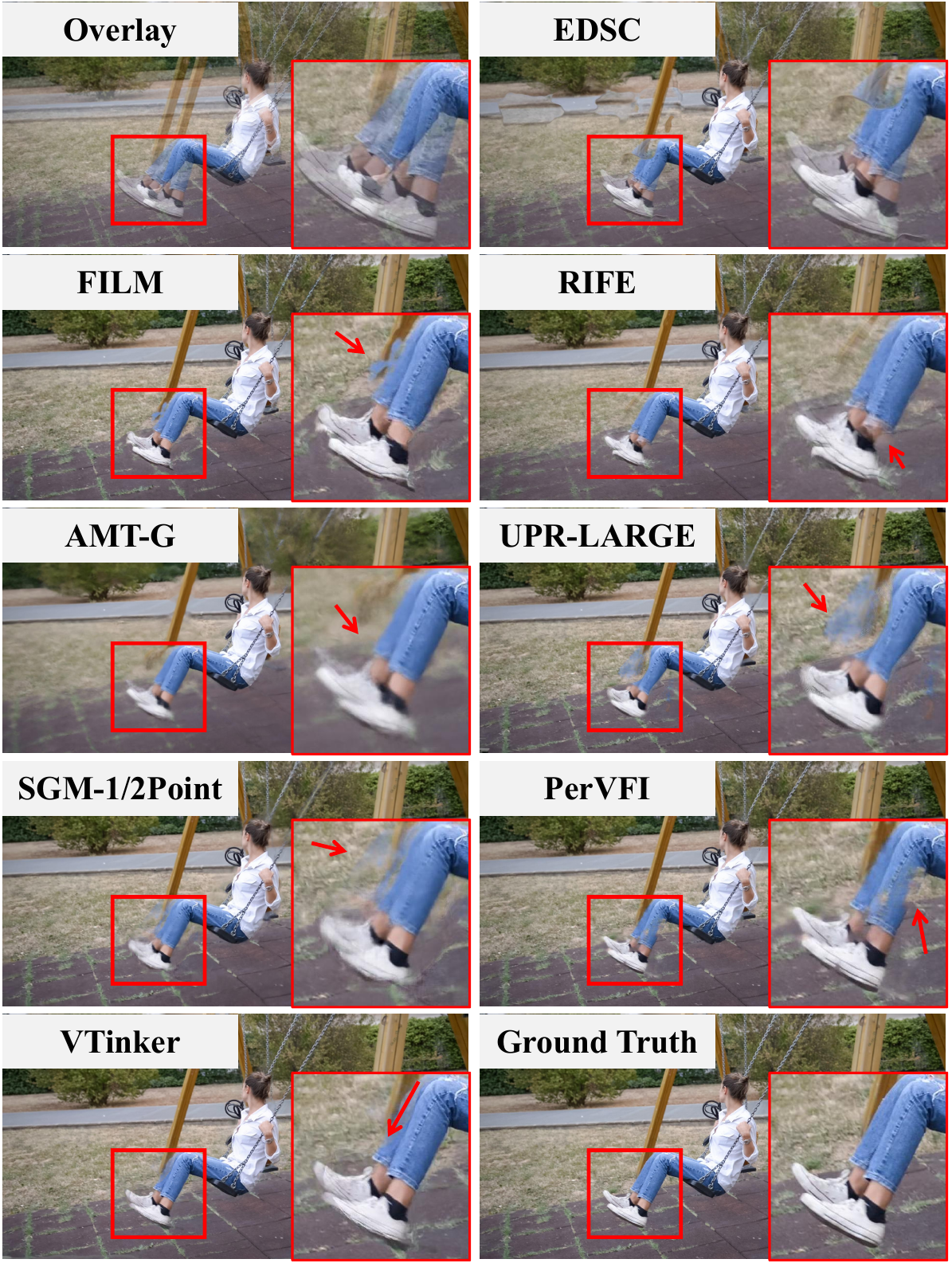}
    \caption{Qualitative comparisons among different methods on 2K resolution. Overlay is the average of two input frames. }
    \label{fig:supvis1}
\end{figure*}

\begin{figure*}[h]
    \centering
    \includegraphics[width=\linewidth, height=1.25\linewidth]{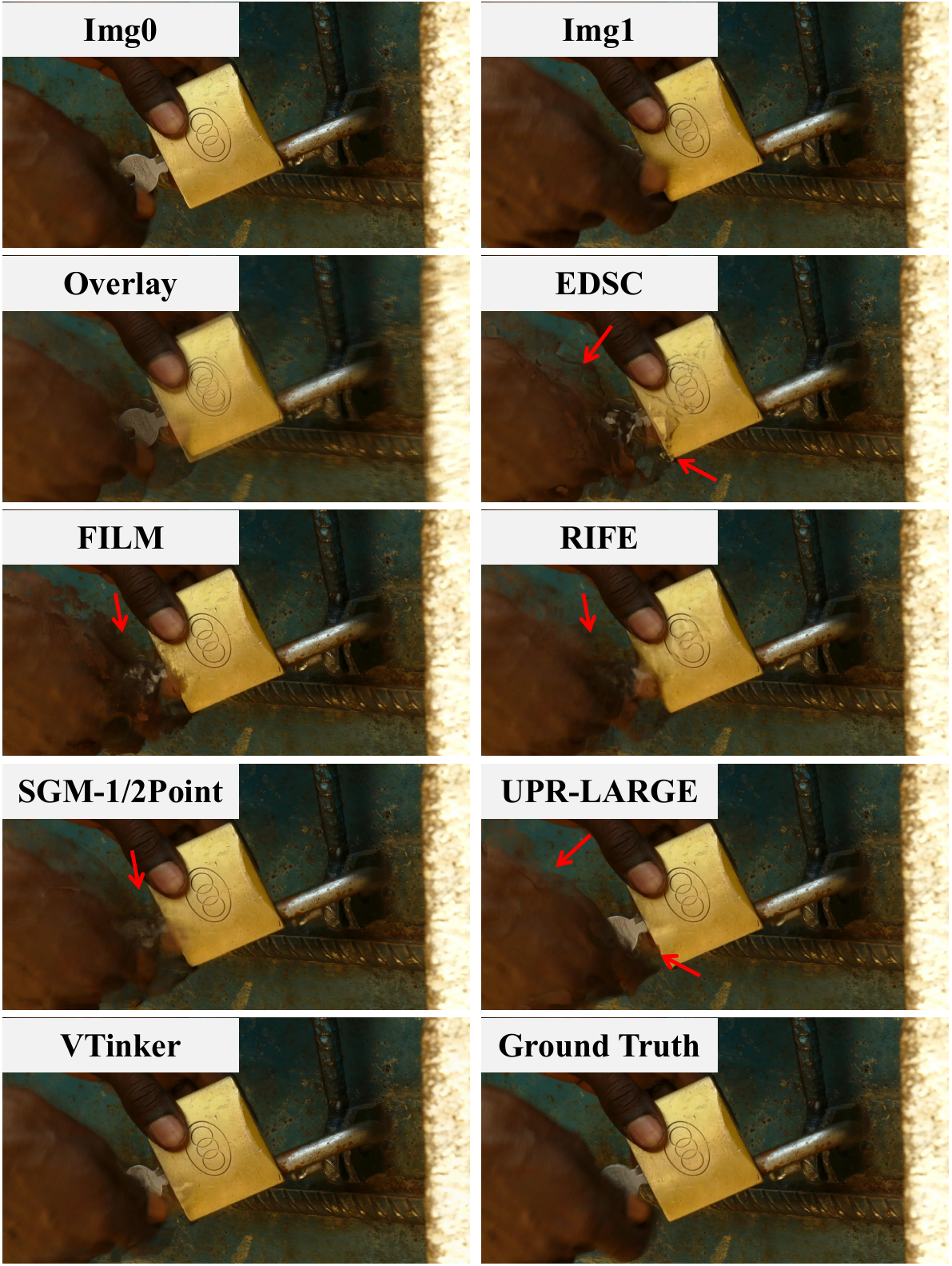}
    \caption{Qualitative comparisons among different methods on 4K resolution. Overlay is the average of two input frames. }
    \label{fig:supvis3}
\end{figure*}

\begin{figure*}[h]
    \centering
    \includegraphics[width=\linewidth, height=1.25\linewidth]{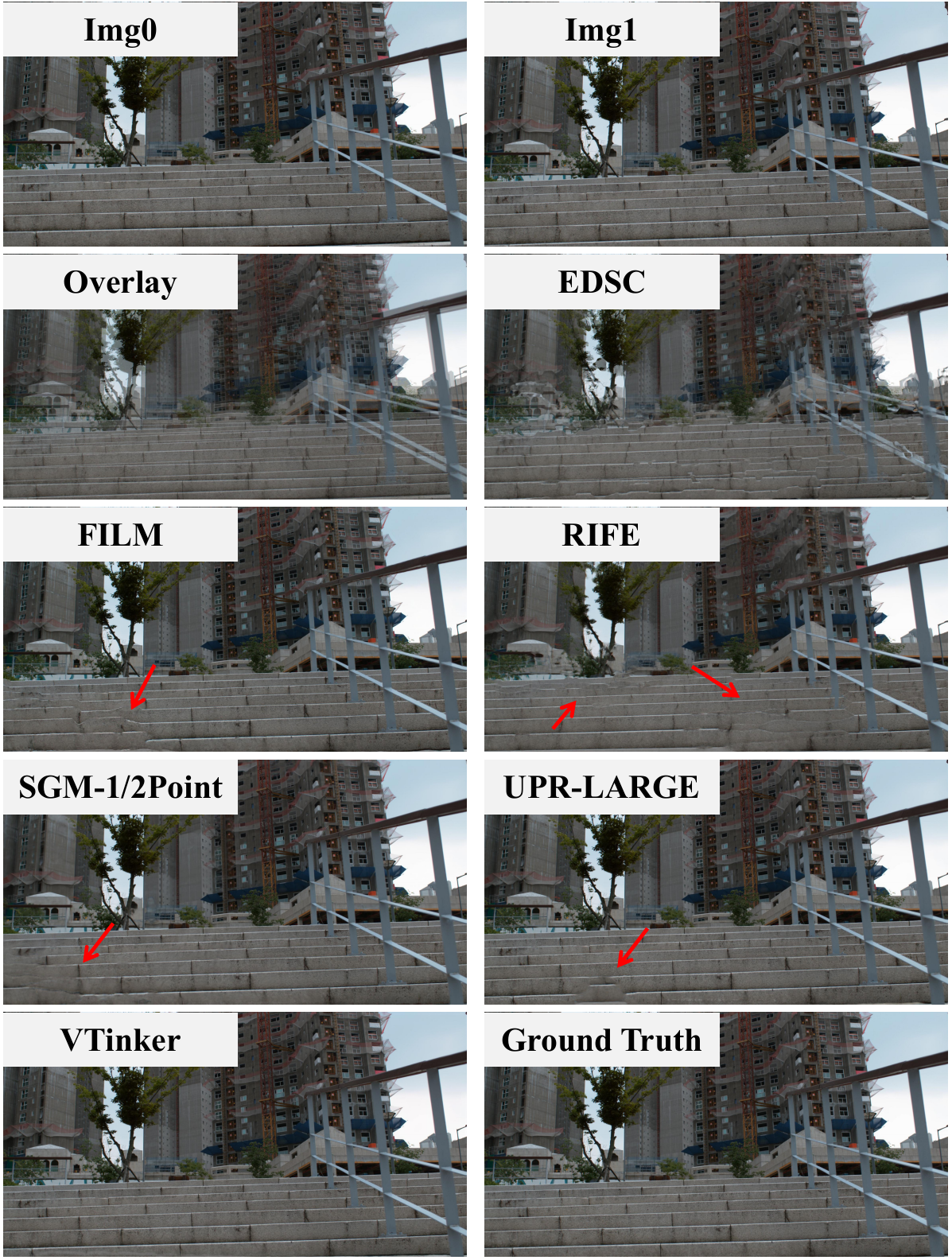}
    \caption{Qualitative comparisons among different methods on 4K resolution. Overlay is the average of two input frames. }
    \label{fig:supvis4}
\end{figure*}

\end{document}